\documentclass[numbers]{article}
\PassOptionsToPackage{numbers}{natbib}

\usepackage[final]{neurips_2025}

\usepackage[utf8]{inputenc} 
\usepackage[T1]{fontenc}    
\usepackage[pagebackref,colorlinks,citecolor=RoyalBlue]{hyperref}       
\usepackage{xurl}            
\usepackage{booktabs}       
\usepackage{amsfonts}       
\usepackage{nicefrac}       
\usepackage{microtype}      
\usepackage{xcolor}         
\usepackage{amsmath}
\usepackage{graphicx}
\usepackage{multirow}
\usepackage{bm}
\usepackage{utfsym}
\usepackage{tcolorbox}
\usepackage[dvipsnames]{xcolor}
\usepackage{cleveref}
\usepackage{subfig}
\usepackage{algorithm}
\usepackage{algpseudocode}
\usepackage{amssymb}

\newcommand{\METHOD}{TPR}
\newcommand{\BfMethod}{\textbf{T}opic-level \textbf{P}reference \textbf{R}ewriting}
\newcommand{\Method}{Topic-level Preference Rewriting}

\newcommand{\ie}{\emph{i.e.}}
\newcommand{\eg}{\emph{e.g.}}

\newcommand{\versus}{\emph{vs.}}

\title{Systematic Reward Gap Optimization for\\Mitigating VLM Hallucinations}

\vspace{-5pt}
\author{
  \vspace{-25pt}\\
  \textbf{Lehan He$^{1,2*}$,\;
  Zeren Chen$^{1,3}$\thanks{Equal contribution. \,\, $^\dag$Corresponding author.}\;\,,\;
  Zhelun Shi$^{1}$,\;
  Tianyu Yu$^4$,\;
  Jing Shao$^{2,3\dag}$,\;
  Lu Sheng$^{1\dag}$} \vspace{3pt} \\
  $^1$School of Software, Beihang University ~~\;\; $^2$Shanghai Innovation Institute \\
  $^3$Shanghai AI Laboratory ~~\;\; $^4$Tsinghua University \vspace{3pt} \\
  \texttt{\small helehan,czr1604,lsheng@buaa.edu.cn, shaojing@pjlab.org.cn} \vspace{6pt}  \\
  Codes and models:~\, \url{https://tpr-dpo.github.io}
  \vspace{-4pt} \\
}

\begin{document}

\maketitle
\begin{abstract}
The success of Direct Preference Optimization (DPO) in mitigating hallucinations in Vision Language Models (VLMs) critically hinges on the true reward gaps within preference pairs.
However, current methods, typically relying on ranking or rewriting strategies, often struggle to optimize these reward gaps in a systematic way during data curation.
A core difficulty lies in precisely characterizing and strategically manipulating the overall reward gap configuration, that is, the deliberate design of how to shape these reward gaps within each preference pair across the data.
To address this, we introduce \Method{} (\METHOD{}), a novel framework designed for the systematic optimization of reward gap configuration.
Through selectively replacing semantic topics within VLM responses with model's own resampled candidates for targeted rewriting, \METHOD{} can provide topic-level control over fine-grained semantic details.
This precise control enables advanced data curation strategies, such as progressively adjusting the difficulty of rejected responses, thereby sculpting an effective reward gap configuration that guides the model to overcome challenging hallucinations.
Comprehensive experiments demonstrate \METHOD{} achieves state-of-the-art performance on multiple hallucination benchmarks, outperforming previous methods by an average of $\sim$20\%.
Notably, it significantly reduces hallucinations by up to 93\% on ObjectHal-Bench, and also exhibits superior data efficiency towards robust and cost-effective VLM alignment.
%
\end{abstract}

\section{Introduction}
\label{sec:intro}

Vision language models (VLMs)~\cite{openai2023gpt4v,liu2023llava,liu2024llavanext,wang2024qwen,lin2024vila} have achieved remarkable success across a spectrum of multimodal tasks, from visual question answering~\cite{wang2024qwen,liu2024llava15} to image captioning~\cite{li2022blip,li2023blip2}, becoming foundational components in modern AI systems.
Despite these advancements, even leading models like GPT-4V~\cite{openai2023gpt4v} suffer from a critical limitation: visual hallucinations~\cite{li2023evaluating,zhou2023analyzing,liu2024survey,yu2024rlhf}.
Specifically, they might confidently describe non-existent objects, misrepresent attributes, or misjudge spatial relationships, contradicting visual inputs.
This poses significant risks, especially in safety-critical scenarios such as autonomous driving~\cite{mao2023gpt} and medicine applications~\cite{thirunavukarasu2023large}.

Recent efforts to mitigate visual hallucinations~\cite{yu2024rlhf,sun2024rlhf,gunjal2024detecting,yu2024rlaifv,jing2024fgaif,zhang2024amp} increasingly leverage preference learning through alignment techniques such as Direct Preference Optimization (DPO)~\cite{rafailov2024direct}.
These methods aim to steer VLM behaviors towards desired outcomes by learning from meticulously curated preference data.
This data typically consists of preference pairs $(y_w, y_l)$ for a given input $x$, where response $y_w$ is preferred over $y_l$.
By aligning with these preference data, the policy learned by DPO implicitly defines a reward function $r(y;x)$ that reflects a probabilistic model like Bradley-Terry $p(y_w\succ y_l|x)$~\cite{bradley1952rank}.
The quality of this learned reward function, as highlighted by recent studies~\cite{zhang2024amp,saeed2024hybrid,wu2024dynamic}, hinges on the fidelity and magnitude of true reward gaps instantiated by each preference pair in the curated data.
Thus, during data curation, strategically designing the reward gaps within each preference pair to shape an effective learning trajectory for the target reward functions, is crucial for robust VLM alignment against hallucinations.
We refer to this deliberate process as the systematic optimization of overall \textbf{\textit{reward gap configuration}}.
Given that these reward gaps are inherently shaped by the intrinsic data characteristics of $y_w$ and $y_l$ (such as informativeness, trustworthiness, and explicit differences), optimizing reward gap configuration necessitates more than mere preference collections, but carefully controlling the underlying data characteristics, thereby sculpting an optimal reward function tailored for minimizing hallucinatory outputs.

\begin{figure}[t]
\begin{center}
\centerline{\includegraphics[width=\linewidth]{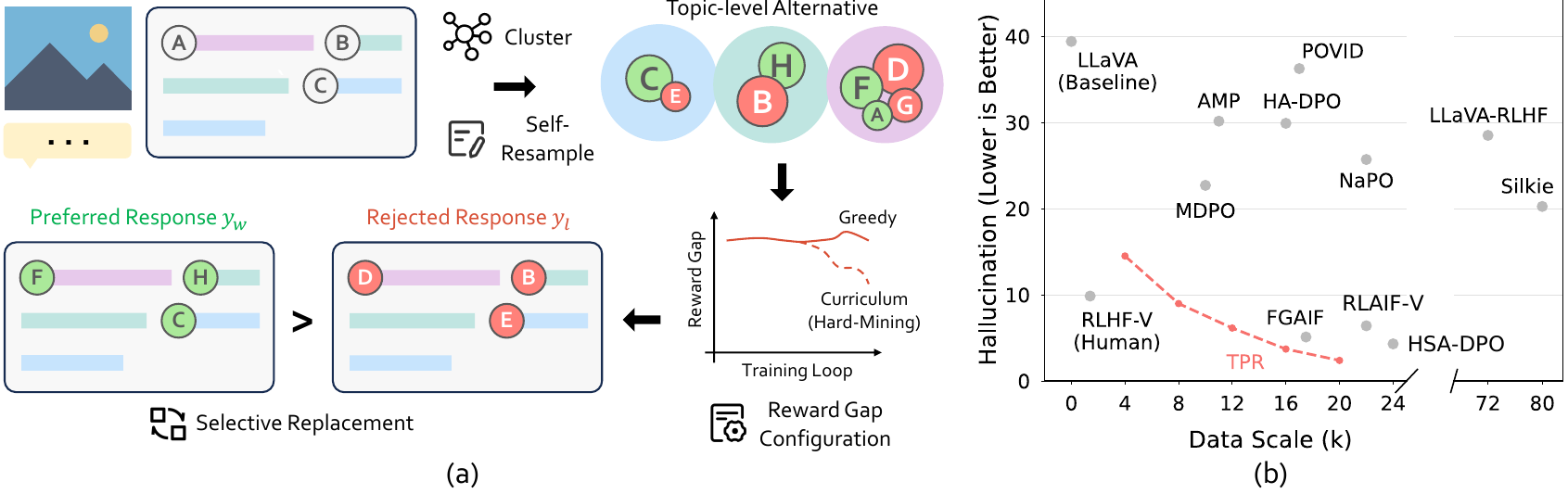}}
\caption{
    \textbf{(a) \Method{}.}
    Based on varying chosen strategies, \METHOD{} selectively replaces each topic using model's internally resampled candidates.
    Here, ``Greedy'' denotes selecting the highest- and worst-scored alternatives for a high-divergence reward gap, while ``Curriculum'' gradually introduces harder-to-discern hallucinations in $y_l$, thereby adjusting the reward gap to master challenging and subtle hallucinations.
    \textbf{(b) Data Efficiency.}
    Apart from manual annotation (RLHF-V~\cite{yu2024rlhf}), \METHOD{} achieves the best data efficiency on visual hallucination reduction.
}
\label{fig:teaser}
\end{center}
\vspace{-3em}
\end{figure}

However, existing methods for preference data curation often lack mechanisms for such deliberate reward gap optimization.
For example, ranking-based methods~\cite{sun2024rlhf,yu2024rlaifv,zhang2024amp,li2024silkie} directly select $y_w$ and $y_l$ from potentially flawed model outputs without correcting underlying hallucinations.
This may lead to low informativeness or insufficient reward gaps to penalize hallucinations, ultimately providing weak signals for learning an effective reward function.
Alternatively, rewriting-based approaches, particularly those~\cite{zhao2023beyond,zhou2024aligning,xiao2024detecting} employing external ``black-box'' models like GPT-4V~\cite{openai2023gpt4v}, encounter difficulties in precisely adjusting the generated responses (\eg, the type and magnitude of the changes) and risk introducing hallucinations in $y_l$ that deviate from model's intrinsic failure modes.
Consequently, both approaches may yield a suboptimal reward gap configuration across the curated data, compromising the learning of robust reward functions.

To address these limitations, we propose \BfMethod{} (\textbf{\METHOD}), a novel VLM hallucination mitigation paradigm designed for systematically optimizing reward gap configuration during data curation, as illustrated in \Cref{fig:teaser} (a).
To provide precise, fine-grained control over semantic details between $y_w$ and $y_l$, \METHOD{} operates at the topic-level.
Specifically, \METHOD{} first decomposes responses into semantic units and clusters them into distinct topics.
It then performs intra-topic self-resampling using the model itself to avoid introducing external biases that may occur with rewriting-based methods.
Preference pairs $(y_w, y_l)$ are subsequently constructed by selectively replacing original semantic units with these alternatives from the same topic.
This mechanism allows for precise, fine-grained adjustment of semantic details between $y_w$ and $y_l$, guided by various chosen strategies.
For example, a greedy strategy can establish highly discriminative reward gaps by constructing $y_w$ and $y_l$ using the highest- or lowest-scored alternatives from their respective topics.

Moreover, this flexible control over fine-grained differences uniquely enables the formulation and investigation of sophisticated strategies for achieving an optimal reward gap configuration.
In this work, we exemplify this capability with a simple yet effective curriculum learning strategy that progressively adjusts the difficulty of hallucinations included in the $y_l$. 
By implementing such hard negative mining, the model can be trained to more effectively counteract subtle and challenging hallucinations, showcasing the effectiveness of \METHOD{}.
Crucially, this effectiveness is coupled with superior data efficiency, stemming directly from \METHOD{}'s ability to curate high-quality preference pairs.
This enhanced data efficiency, clearly depicted in \Cref{fig:teaser} (b), underscores how \METHOD{}'s systematic optimization of reward gap configuration leads to more performant and cost-effective VLM alignment.

In summary, our contributions can be summarized as below:
\begin{itemize}

    \item We underscore the importance of systematically optimizing reward gap configuration during data curation for robust VLM alignment, an aspect often overlooked in prior studies.

    \item We propose \Method{} (\METHOD{}), a novel paradigm designed to offer fine-grained control over individual reward gaps during data curation, achieved through topic-level selective replacement with the model's own resampled candidates.

    \item Benefiting from the fine-grained control afforded by \METHOD{}, we introduce a curriculum learning strategy that optimizes the overall reward gap configuration by progressively adjusting the difficulty of rejected responses $y_l$, enhancing robustness against challenging hallucinations.

    \item Comprehensive experiments demonstrate that \METHOD{} achieves the state-of-the-art performance on multiple visual hallucination benchmarks, outperforming previous methods in both performance (by $\sim$20\%) and data efficiency.
    
\end{itemize}

\section{Related Work}
\label{sec:related_work}

\textbf{Vision Language Models and Hallucinations.}
The advent of Large Language Models (LLMs)~\cite{openai2023gpt4,touvron2023llama2,liu2024deepseek,yang2024qwen} has driven the development of Vision Language Models (VLMs)~\cite{openai2023gpt4v,liu2023llava,wang2024qwen,lin2024vila}.
Through multimodal alignment followed by supervised instruction tuning, VLMs achieve remarkable proficiency in visual perception and comprehensive understanding.
However, their tendency to produce hallucinations~\cite{li2023evaluating,zhou2023analyzing,liu2024survey,bai2024hallucination}, \ie, generating responses that not factually grounded in the given image, undermines their reliability and practicality in real-world applications.
These hallucinations can be attributed to multiple factors, including inherent biases inherited from LLMs~\cite{huang2023survey}, biased training data~\cite{chuang2023debiasing,tu2023many}, insufficient multimodal alignment~\cite{fu2023mme} and suboptimal inference strategies~\cite{huang2024opera}.
Efforts to mitigate hallucinations broadly fall into training-free and training-based approaches.

\textbf{Training-free Hallucination Reduction.}
Training-free approaches~\cite{zhai2023halle,zhao2024marine,yin2023woodpecker,zou2024look,wang2024mllm} aim to reduce hallucinations without additional model training, typically by intervening during the inference stage.
These approaches often involve adjusting decoding strategies or implementing post-hoc output correction mechanisms.
For example, HallE-Switch~\cite{zhai2023halle} modifies the decoding process to suppress object predictions with low confidence scores. 
MARINE~\cite{zhao2024marine} employs classifier-free guidance using auxiliary object grounding features to enrich the visual context during generation. 
Woodpecker~\cite{yin2023woodpecker} operates post-hoc, identifying and rectifying factual inconsistencies in generated text by leveraging feedback from a more capable VLM. 
While efficient, these methods primarily address the symptoms rather than the core deficiencies of hallucinations within the model itself and may offer limited improvements against deeply ingrained hallucinatory tendencies.

\textbf{Training-based Hallucination Reduction.}
Training-based methods~\cite{yu2024rlhf,sun2024rlhf,gunjal2024detecting,yu2024rlaifv,jing2024fgaif,zhang2024amp,xiao2024detecting,liu2023mitigating} predominantly learning from preference through alignment techniques such as Direct Preference Optimization (DPO)~\cite{rafailov2024direct}.
Preference data curation is central to these methods, as the policy's learned behavior depends significantly on the quality of the training supervision embedded in the preference data.
To construct high-quality preference data, these methods typically adopt ranking-based or rewriting-based strategies.
Ranking-based methods~\cite{yu2024rlaifv,zhang2024amp,li2024silkie} often employ an auxiliary labeler model to distinguish preferred responses.
For example, RLAIF-V~\cite{yu2024rlaifv} implements a divide-and-conquer strategy that aggregates scores from decomposed sub-responses, reducing dependence on proprietary models.
AMP~\cite{zhang2024amp} constructs multi-level preferences by contrasting outputs from models of varying scales, enabling cross-level comparison.
On the other hand, rewriting-based methods involve modifying an initial response to create a preferred response $y_w$ or a rejected response $y_l$. 
This rewriting can be performed manually by human annotators~\cite{yu2024rlhf,gunjal2024detecting} or automated using AI rewriters~\cite{zhao2023beyond,zhou2024aligning,xiao2024detecting} like GPT-4V~\cite{openai2023gpt4v}.
Exploring effective data curation approaches remains an active area of research, motivating the development like \METHOD{} proposed in this work.

\section{\Method{}}
\label{sec:method}

\subsection{Preliminary: Preference Data and Alignment}
\label{subsec:preliminary}

Mitigating visual hallucinations in VLMs often involves aligning a policy model $\pi_\theta$ with carefully curated preference data, denoted as $\mathcal{D}=\{(I, x,y_w,y_l)\}$.
Typically, for a given image $I$ and prompt $x$, these methods employ a base reference model $\pi_\text{ref}$ to generate candidate responses. 
Human experts or proxy AI labelers $\pi_\text{label}$ then evaluate these responses or rewrite them to form preference pairs $(y_w, y_l)$, where $y_w$ is preferred over $y_l$.
The target policy model $\pi_\theta$, often initialized from $\pi_\text{ref}$, is subsequently fine-tuned on $\mathcal{D}$ using Direct Preference Optimization (DPO)~\cite{rafailov2024direct}.

The core idea of DPO is to optimize the policy $\pi_\theta$ to satisfy the preferences in $\mathcal{D}$, which are assumed to follow a latent reward model that reflects the preference probability $p(y_w\succ y_l|x)$, while simultaneously being constrained by a KL-divergence penalty to not stray too far from the initial reference policy $\pi_\text{ref}$.
The DPO loss function is formulated as:
\begin{equation}
    \mathcal{L}_{\text{DPO}}(\pi_\theta; \pi_{\text{ref}}) = -\mathbb{E}_{(x, y_w, y_l) \sim \mathcal{D}} \left[ \log\sigma\left(\beta \log\frac{\pi_\theta(y_w|x)}{\pi_{\text{ref}}(y_w|x)} - \beta \log\frac{\pi_\theta(y_l|x)}{\pi_{\text{ref}}(y_l|x)}\right) \right]
\end{equation}
where $\beta$ is a hyperparameter that controls the strength of the preference modeling versus the KL constraint.

As highlighted by recent studies~\citep{zhang2024amp,saeed2024hybrid,wu2024dynamic}, the effectiveness of such alignment critically depends on the true reward gaps, formally defined as the difference $r(y_w;x) - r(y_l;x)$ within each preference pair.
The strategic configuration of reward gaps across the curated preference data provides substantial training signals that not only reinforce desired behaviors but also accurately expose the model's genuine and harder-to-discern deficiencies.
Therefore, to sculpt effective reward gaps, two key principles should guide data curation:
\textbf{(1)} The process must enable fine-grained control over data characteristics in $y_w$ and $y_l$ to deliberately shape reward gaps.
\textbf{(2)} The preference pairs should accurately reflect desired behaviors while exposing model's intrinsic failure modes.

Guided by these principles, we propose \Method{} (\METHOD{}), a novel framework designed to systematically configure the reward gap by establishing fine-grained control over topic-level semantic details within the response.
\METHOD{} implements this through two core steps: topic-level alternatives generation (\Cref{subsec:topic_alter_gen}) and selective topic replacement (\Cref{subsec:selective_topic_rewrite}).
Building upon this precise control, \METHOD{} enables the active exploration and strategic shaping of reward gap configuration, exemplified by the curriculum learning strategy detailed in \Cref{subsec:hard_negative_mine}.
For clarity, we provide a detailed pseudo code of the complete \METHOD{} workflow in~\Cref{alg:tpr}.

\begin{algorithm}[H]
    \caption{Topic-level Preference Rewriting (TPR)}
    \label{alg:tpr}
    \begin{algorithmic}[1] 
        \Require Reference model $\pi_\text{ref}$, Labeler model $\pi_\text{label}$, Source data $\mathcal{D}_\text{src}$ (Image $I$, Prompt $x$), Chosen strategy $\omega$ (e.g., greedy, curriculum).
        \Ensure Preference data $\mathcal{D}_\text{pref} = \{(I, x, y_w, y_l)\}$.
        \State Initialize $\mathcal{D}_\text{pref} \leftarrow \varnothing$;
        \For{each ($I$, $x$) in $\mathcal{D}_\text{src}$}
            \State Initialize initial responses $S_y \leftarrow \varnothing$, semantic units $S_u \leftarrow \varnothing$, topic clusters $S_C \leftarrow \varnothing$;
            \For{$i \leftarrow 1$ to $M$}
                \State $y_i \leftarrow \text{Sample}(\pi_\text{ref}, I, x)$;
                \State Add $y_i$ to $S_y$;
                \State Add $\text{Decompose}(\pi_\text{ref}, y_i)$ to $S_u$;
            \EndFor
            \State $S_C \leftarrow \text{TopicCluster}(\pi_\text{ref}, S_u)$;
            \For{each cluster $C$ in $S_C$}
                \State $\text{IntraTopicResample}(\pi_\text{ref}, C)$;
                \State $\text{Rank}(\pi_\text{label}, C)$;
            \EndFor
            \State Initialize response template $y_k \leftarrow$ Randomly select from $S_y$;
            \State Initialize replacements $S_w \leftarrow \varnothing, S_l \leftarrow \varnothing$;
            \For{each unit $u_k \in C$ in $y_k$}
                \State $(S_w, S_l) \leftarrow \text{SelectAlternatives}(\omega, C)$;
            \EndFor
            \State $y_w \leftarrow \text{InContextRewrite}(\pi_\text{ref}, y_k, S_w)$;
            \State $y_l \leftarrow \text{InContextRewrite}(\pi_\text{ref}, y_k, S_l)$;
            \State Add ($I$, $x$, $y_w$, $y_l$) to $\mathcal{D}_\text{pref}$;
        \EndFor
        \State \Return $\mathcal{D}_\text{pref}$
    \end{algorithmic}
\end{algorithm}

\subsection{Topic-level Alternatives Generation}
\label{subsec:topic_alter_gen}

\begin{figure}[t]
\centering
\includegraphics[width=\linewidth]{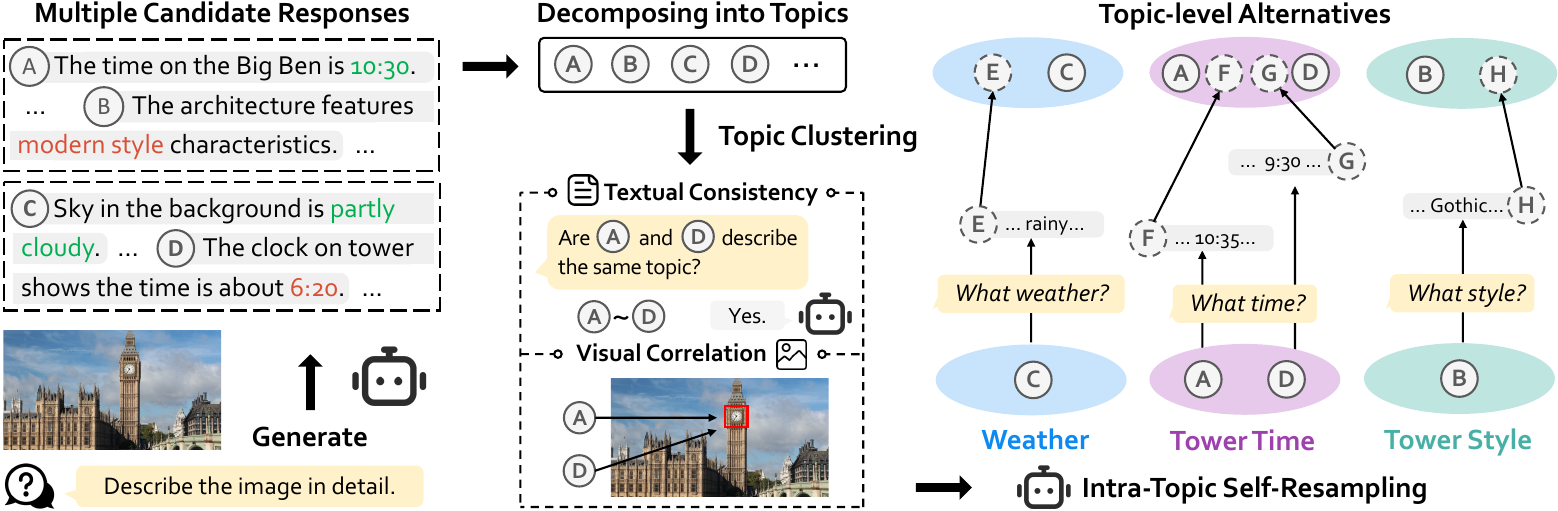}
\caption{
    \textbf{Obtaining High-quality and Diverse Topic-level Alternatives.}
    Initially, candidate responses from VLM are decomposed into fine-grained semantic units. 
    These units are then grouped into distinct topic clusters based on textual consistency and visual correlation.
    A diverse pool of topic-level alternatives is then generated via intra-topic self-resampling.
}
\label{fig:topic_alter_gen}
\vspace{-1em}
\end{figure}

VLM responses comprise various semantic topics, encompassing diverse objects and attributes, intricate spatial relationships, or subtle contextual implications.
To offer flexible control over fine-grained details within these responses, \METHOD{} operates precisely at the topic level.
Such a topic-centric approach is made viable by findings like those in RLAIF-V~\cite{yu2024rlaifv}, which suggest topics within a response often exhibit weak correlations, permitting their relatively independent manipulation.
As illustrated in \Cref{fig:topic_alter_gen}, \METHOD{} introduces a topic-level alternative generation approach designed to provide a rich set of candidates for the subsequent selective replacement of individual semantic topics.

\textbf{Decomposing into Topics.}
Following established protocol~\cite{yu2024rlhf,yu2024rlaifv}, for an input image $I$ and prompt $x$, we begin by sampling multiple candidate responses $\{y_1,\dots,y_M\}$ from the reference model $\pi_\text{ref}$.
We then utilize $\pi_\text{ref}$ to decompose each candidate response $y_m$ into a set of fine-grained semantic units $\{u_{m,1},\dots,u_{m,N_m}\}$, where each semantic unit $u_{m,n}$ corresponds to a distinct semantic topic.

\textbf{Topic Clustering.}
To facilitate subsequent selective replacement of semantic units with alternatives under the same topic, we group semantically related units from different candidate responses into unique topic clusters, based on textual consistency and visual correlation.
Textual consistency between units $u_{m,n}$ and $u_{p,q}$ is assessed by querying the reference model $\pi_\text{ref}$, \eg, ``Are $u_{m,n}$ and $u_{p,q}$ describing the same topic?''.
For visual correlation, we utilize features from the VLM's visual encoder (\eg, CLIP~\cite{radford2021learning}) to verify whether both units refer to similar regions within the input image $I$.
It is essential for disambiguating textually similar units that describe visually distinct entities.
Units $u_{m,n}$ and $u_{p,q}$ are considered the same topic only if they satisfy both textual consistency and visual correlation criteria.
We then apply a greedy algorithm~\cite{blondel2008fast} for topic clustering, yielding topic cluster $\{c\}$, each containing textually and visually related units.
More details are provided in Appendix~\ref{sec:appendix_topic_cluster}.

\textbf{Intra-Topic Self-Resampling.}
To enrich the pool of alternatives within each topic cluster, we employ the reference model $\pi_\text{ref}$ to perform self-resampling focused on individual topics. 
Specifically, for each decomposed unit $u_{m, n}$, we prompt $\pi_\text{ref}$ to first convert the unit into a relevant wh-question (\eg, \textit{``The time on the Big Ben is 3:30.''} $\rightarrow$ \textit{``What time is on the Big Ben?''}).
We then query $\pi_\text{ref}$ multiple times with these topic-specific questions to obtain a set of candidate semantic units pertinent to that topic, which avoids introducing potential biases or hallucinations from external models.
Compared to resampling entire responses, this intra-topic self-resampling offers two key advantages central to \METHOD{}.
First, by focusing $\pi_\text{ref}$ on one topic at a time, it boosts the efficiency of obtaining valid and diverse semantic units, bypassing the demands of simultaneous correctness across all units required in entire response resampling.
Second, generating alternatives at the topic level provides the necessary granularity for the subsequent selective replacement, enabling systematic shaping of the desired reward gaps in the resulting preference pairs.

\subsection{Selective Topic Rewriting}
\label{subsec:selective_topic_rewrite}

\textbf{Intra-Topic Ranking.}
To distinguish between accurate and potentially hallucinatory semantic units within each topic cluster $c$, we adopt an intra-topic ranking strategy.
For each semantic unit $u_{m,n}^c$ (including original and self-resampled alternatives), we prompt the $\pi_\text{ref}$ to convert it into a corresponding yes-no question (\emph{e.g.}, \textit{``The time on the Big Ben is 3:30.''} $\rightarrow$ \textit{``Is the time on the Big Ben 3:30?''}).
Given the input image $I$, we then query $\pi_\text{label}$ with converted yes-no questions, obtaining probabilities $p_\text{Y}$ and $p_\text{N}$ for ``Yes'' or ``No'' responses, respectively.
Each unit $u_{m,n}^c$ is assigned a score $S(u_{m,n}^c) = p_\text{Y} - p_\text{N}$, where a higher score indicates a higher likelihood that the unit is factually accurate and non-hallucinatory for topic $c$..
Moreover, as noted in prior work~\cite{yu2024rlaifv}, evaluating fine-grained units tends to yield more reliable assessments, allowing even moderately capable models (\eg, $\pi_\text{ref}$ itself) to serve effectively as $\pi_\text{label}$ in \METHOD{}.

\begin{figure}[t]
\centering
\includegraphics[width=\linewidth]{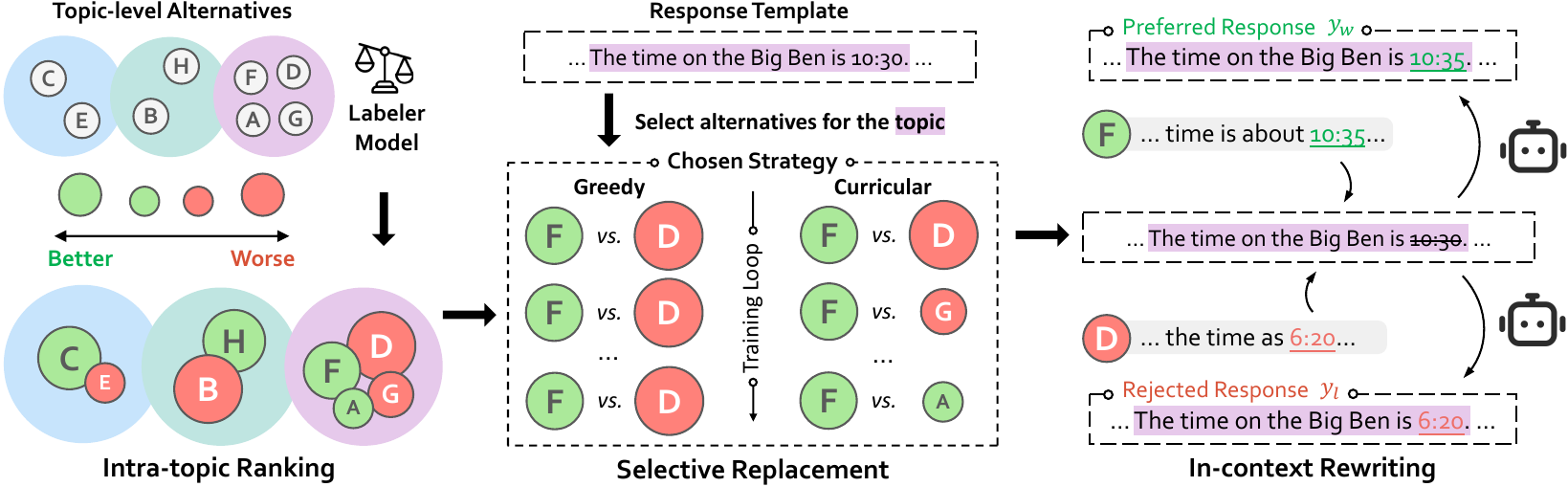}
\caption{
    \textbf{Constructing Preference Pairs by Selectively Replacement.}
    Starting with a pool of scored topic alternatives, preference pairs are constructed through selectively replacing units within the template.
    The specific alternatives are chosen based on strategies like greedy replacement, to deliberately control over the resulting reward gap.
    In-context rewriting is then employed to seamlessly integrate these chosen alternatives, ensuring fluent preference pairs for subsequent model alignment.
}
\label{fig:selective_topic_rewrite}
\vspace{-1em}
\end{figure}

\textbf{Selective Replacement.}
Leveraging the ranked semantic units within each topic, we construct preference pairs $(y_w, y_l)$ through a selective replacement mechanism.
First, we randomly select one of the initial candidate responses $\{y_1, y_2, \dots, y_M\}$ as a response template, denoted $y_k$.
This template consists of its original semantic units $\{u_{k,1}^{c_1},u_{k,2}^{c_2},\dots\}$, each associated with a topic.
We then generate $y_w$ and $y_l$ by selectively replacing units $u_{k,n}^{c_n}$ with alternative units chosen from the ranked pool within the same topic cluster $c_n$.
This selective replacement is the core mechanism for providing fine-grained control over the semantic difference between $y_w$ and $y_l$.
For example, a common strategy aimed at constructing highly discriminative pairs involves replacing each $u_{k,n}^{c_n}$ within the template with the highest- and lowest-scoring alternatives from their respective topic clusters.
By varying chosen strategies for replacement units, this mechanism allows for deliberate adjustment of each semantic unit, consequently, the reward gap encoded within each $(y_w, y_l)$ pair.

\textbf{In-Context Rewriting.}
Directly substituting alternative units into the template could potentially disrupt the natural language flow and stylistic consistency, producing awkward or incoherent responses.
To mitigate this issue, we perform the selective replacement using in-context rewriting guided by the reference model $\pi_\text{ref}$.
Specifically, we instruct $\pi_\text{ref}$ to integrate the semantic content of the chosen replacement units into the template response, preserving logical structure and linguistic style.
This results in preference pairs $(y_w,y_l)$ where the semantic differences related to specific topics are precisely controlled while maintaining overall response quality. 
These high-quality preference pairs, encoding intentionally designed reward gaps, are then used to fine-tune the policy model $\pi_\theta$ via DPO, effectively steering it away from generating visual hallucinations.

\subsection{Exploring Optimal Reward Gap Configuration}
\label{subsec:hard_negative_mine}

Constructing preference pairs $(y_w, y_l)$ where corresponding semantic units exhibit maximal score divergence provides a strong initial signal in DPO training.
However, recent studies~\cite{zhang2024amp,wu2024dynamic} suggest such a greedy strategy may not yield the optimal reward gap configuration for robust alignment.
Focusing exclusively on maximizing individual reward gaps might prioritize penalizing obvious, easily detectable errors over the subtle, challenging hallucinations that the model genuinely struggles with, potentially leading to inefficient learning.
The systematic control over fine-grained semantic difference afforded by \METHOD{} allows us to explicitly optimize the reward gaps embedded in the curated data, thereby refining the model's ability to avoid these harder-to-discern failure modes.

Leveraging this control, we propose a simple yet effective curriculum learning strategy, analogous to hard negative mining in other domains, to optimize the reward gap configuration for hallucination reduction.
This strategy involves structuring the preference data curation and DPO training in stages.
Specifically, during an initial ``\textbf{Warm-Up}'' stage, we adopt the greedy strategy described above, constructing $(y_w, y_l)$ pairs with maximal score divergence between corresponding semantic units.
This provide a strong initial learning signal, encouraging faster convergence by clearly differentiating between highly discriminative content.
In a subsequent ``\textbf{Hard-Mining}'' stage, we gradually increase the difficulty of the learning objective by constructing $y_l$ using incorrect alternative units that have progressively higher scores, \ie, they are less obviously wrong and closer to the decision boundary.
Introducing these ``hard negative'' examples in $y_l$ challenges the model to make finer distinctions during DPO training. 
By learning to differentiate preferred responses from subtly flawed ones, the model can more effectively refine its ability to detect and avoid nuanced hallucinations, ultimately improving robustness. 
This curriculum strategy systematically shape the overall reward gap configuration over time, starting with broad distinctions and moving towards finer-grained ones ultimately.
Further details are provided in Appendix~\ref{sec:appendix_curriculum_learning}.

\section{Experiments}
\label{sec:exp}

\subsection{Experimental Setup}
\label{subsec:exp_setup}

\textbf{Models.}
In line with previous studies, we use LLaVA-1.5-7B~\cite{liu2024llava15} as both the reference model $\pi_\text{ref}$ for generating preference data, and as the policy model $\pi_\theta$ that is subsequently fine-tuned.
The labeler model $\pi_\text{label}$, used for intra-topic ranking described in \Cref{subsec:selective_topic_rewrite}, is LLaVA-NeXT-34B~\cite{liu2024llavanext}, a choice consistent with recent approaches~\cite{yu2024rlaifv,xiao2024detecting}.
In our experiments, we evaluate two variants of proposed \METHOD{}: 
(1) \textbf{\METHOD{}} utilizes the greedy replacement strategy for data curation.
(2) \textbf{\METHOD{}-CL} (\textbf{C}urriculum \textbf{L}earning) utilizes a curriculum learning strategy for data curation, detailed in \Cref{subsec:hard_negative_mine}.
We compare \METHOD{} against a range of counterparts, including ranking-based methods~\cite{sun2024rlhf,yu2024rlaifv,jing2024fgaif,zhang2024amp,li2024silkie} and rewriting-based methods that rely on human experts~\cite{yu2024rlhf} or external models~\cite{zhou2024aligning,xiao2024detecting,zhao2023beyond,yang2025mitigating}.

\begin{table}[t]
\caption{
    \textbf{
        Experimental Results on Several Hallucinations and General Capabilities Benchmarks.
    }
    The best and second best results are shown in \textbf{bold} and \underline{underlined}, respectively.
}
\label{tab:main_results}
\begin{center}
\resizebox{\linewidth}{!}{%
\begin{tabular}{@{}l|cccccccccc|cc}
    \toprule
    & 
    \multicolumn{10}{c|}{\textbf{Hallucination Benchmarks}} & 
    \multicolumn{2}{c}{\textbf{General Benchmarks}} \\
    \cmidrule(lr){2-11} \cmidrule(lr){12-13}
    \multirow{3}{*}{\vspace{+8mm}\textbf{Model}} & 
    \multicolumn{2}{c}{\textbf{ObjHal}} &
    \multicolumn{2}{c}{\textbf{MMHal}} &
    \multicolumn{2}{c}{\textbf{AMBER}} &
    \multicolumn{2}{c}{\textbf{POPE}} &
    \multicolumn{2}{c|}{\textbf{RefoMB}} &
    \multicolumn{1}{c}{\textbf{LLaVA-B}} &
    \multicolumn{1}{c}{\textbf{MMstar}} \\ 
    \cmidrule(lr){2-3} \cmidrule(lr){4-5} \cmidrule(lr){6-7} \cmidrule(lr){8-9} \cmidrule{10-11} \cmidrule(lr){12-12} \cmidrule(lr){13-13} 
    &
    CH$_\text{s}$ $\downarrow$ & CH$_\text{i}$ $\downarrow$ &
    Score $\uparrow$ & Hall. $\downarrow$ &
    Acc. $\uparrow$ & F1 $\uparrow$ &
    Adv. $\uparrow$ & All $\uparrow$ &
    Trust. $\uparrow$ & Win. $\uparrow$ &
    Acc. $\uparrow$ & 
    Acc. $\uparrow$ \\
    \midrule
    LLaVA-RLHF-13B~\cite{sun2024rlhf}         & 38.1 & 18.9 & 2.02 & 62.5 & 79.7 & 83.9 & 82.3 & 81.9 & 26.3 & 17.2 & 61.5 & \underline{34.2} \\
    RLHF-V-13B~\cite{yu2024rlhf}              & 12.2 & 7.5  & 2.45 & 51.0 & 72.6 & 75.0 & 80.5 & 81.9 & 41.4 & 17.7 & 51.4 & 33.2 \\
    Silkie-10B~\cite{li2024silkie}            & 27.1 & 13.4 & \textbf{3.19} & 32.3 & 82.2 & \underline{87.6} & 80.3 & 81.1 & 38.9 & 21.2 & \textbf{73.2} & 33.6 \\
    POVID-7B~\cite{zhou2024aligning}          & 48.1 & 24.4 & 2.08 & 56.2 & \textbf{82.9} & 87.4 & 84.0 & 85.8 & 44.4 & 13.6 & 62.2 & \textbf{34.3} \\
    MFPO-7B~\cite{jiang2024modality}          & 13.4 & 6.6  & 2.69 & 49.0 & -- & -- & -- & -- & -- & -- & -- & -- \\
    HA-DPO-7B~\cite{zhao2023beyond}           & 39.9 & 19.9 & 1.98 & 60.4 & 75.2 & 79.9 & 82.5 & 86.9 & 39.9 & 17.2 & 67.2 & 32.9 \\
    OPA-DPO-7B~\cite{yang2025mitigating}      & 13.0 & 4.3 & 2.83 & 45.8 & 81.3 & 85.6 & 83.7 & 86.1 & 39.4 & 18.2 & 62.2 & 32.2 \\
    mDPO-7B~\cite{wang2024mdpo}                  & 35.7 & 9.8 & 2.39 & 54.2 & -- & -- & -- & -- & -- & -- & -- & -- \\
    AMP-MEG-7B~\cite{zhang2024amp}            & 37.8 & 22.5 & \underline{3.17} & 35.0 & 78.3 & 83.6 & 83.4 & 86.8 & 42.9 & 18.7 & 54.6 & 27.5 \\
    RLAIF-V-7B~\cite{yu2024rlaifv}            & 8.5  & 4.3  & 3.06 & \textbf{29.2} & 76.8 & 84.5 & 81.2 & 83.3 & 47.5 & 20.7 & 64.9 & 31.8 \\
    FGAIF-7B~\cite{jing2024fgaif}                & 6.2  & 3.9  & 3.09 & 36.0 & -- & -- & 79.9 & 83.4 & -- & -- & -- & -- \\
    HSA-DPO-13B~\cite{xiao2024detecting}      & 5.3  & 3.2  & 2.61 & 48.0 & --   & --   & -- & -- & -- & -- & -- & --   \\
    \noalign{\smallskip} \hline \noalign{\smallskip}
    LLaVA-1.5-7B~\cite{liu2024llava15}        & 53.6 & 25.2 & 2.36 & 51.0 & 73.5 & 77.6 & \textbf{84.5} & 85.9 & 30.8 & 12.1 & 59.7 & 30.3 \\ 
    \;\;+\;\METHOD{}-7B                       & \underline{4.0} & \underline{2.2}  & 3.01 & 31.2 & 82.3 & \underline{87.6} & 83.5 & 86.2 & \underline{58.1} & \underline{31.3} & 69.2 & 33.2 \\
    \;\;+\;\METHOD{}-CL-7B                    & \textbf{3.4} & \textbf{1.8}  & 3.06 & \underline{30.2} & \underline{82.7} & \textbf{87.8} & \underline{84.2} & \textbf{87.6} & \textbf{61.1} & \textbf{32.3} & \underline{71.1} & 33.3 \\
    \bottomrule
\end{tabular}%
}
\end{center}
\vspace{-1em}
\end{table}

\textbf{Data Source.}
Following~\citet{yu2024rlhf,yu2024rlaifv}, we curate preference data based on 7 publicly available dataset sources: VQA v2~\cite{goyal2017vqav2}, MSCOCO~\cite{lin2014mscoco}, ShareGPT-4V~\cite{chen2023sharegpt4v}, TextVQA~\cite{singh2019textvqa}, MovieNet~\cite{huang2020movienet}, OKVQA~\cite{marino2019okvqa} and Google Landmark v2~\cite{weyand2020googlelandmarkv2}.
We generate a total of 20,000 preference data instances used for alignment.
For the \METHOD{}-CL variant, 12,000 instances (60\%) are constructed during ``Warm-Up'' stage, and the remaining 8,000 (40\%) are constructed during ``Hard-Mining'' stage.

\textbf{Preference Learning.}
We apply Direct Preference Optimization (DPO)~\cite{rafailov2024direct} for preference learning, aligning the policy model with the preference data curated by \METHOD{} and its variants.
We use the AdamW~\cite{loshchilov2017adamw} optimizer with a batch size of 8, a learning rate of $5{\times}10^{-7}$ with the cosine decay strategy.
The policy model is fine-tuned for 1 epoch on 8 NVIDIA A100 GPUs.

\textbf{Evaluation Benchmarks.}
We assess visual hallucinations mitigation on Object HalBench~\cite{rohrbach2018objhal}, MMHal-Bench~\cite{sun2024rlhf}, AMBER~\cite{wang2023amber} (discriminative part), RefoMB~\cite{yu2024rlaifv} and POPE~\cite{li2023evaluating}.
We assess general capabilities on LLaVA-Bench~\cite{liu2023llava} (in-the-wild) and MMStar~\cite{chen2024mmstar}.

\subsection{Main Results}
\label{subsec:main_results}

Results in \Cref{tab:main_results} demonstrate \METHOD{} significantly mitigates visual hallucinations in VLMs across several benchmarks.
The proposed \METHOD{}, in both its evaluated variants, achieves leading performance in mitigating visual hallucinations when applied to LLaVA-1.5-7B.
Notably, our \METHOD{}-CL variant establishes new state-of-the-art results on several hallucination benchmarks, reducing hallucinations by $\sim$93\% on Object-HalBench and $\sim$41\% on MMHal-Bench.
This substantial improvements can be consistently observed across other challenging hallucination benchmarks such as AMBER and RefoMB, underscoring the efficacy of our proposed design.
Moreover, models fine-tuned using \METHOD{}-curated data not only maintain but sometimes enhance the base model's performance on general capability benchmarks.
This outcome indicates that \METHOD{} can effectively suppresses visual hallucinations without compromising general VLM capabilities.

The introduction of a curriculum learning strategy (\METHOD{}-CL) consistently yields superior performance compared to the greedy variant across all evaluated metrics.
By progressively exposing the model to more difficult examples according to a curricular scheduler, \METHOD{}-CL compels it to discern fine-grained details, complex contextual relationships, and subtle inconsistencies that often underpin persistent and challenging visual hallucinations.
Crucially, \METHOD{}'s unique capability to exercise fine-grained control over reward gap configuration is the key to effectively implementing curriculum learning strategy, thereby cultivating a more robust policy and delivering the state-of-the-art hallucination mitigation demonstrated by \METHOD{}-CL.

\newcommand{\diff}[1]{\textbf{\textcolor{Periwinkle}{#1}}}

\begin{table}[t]
\caption{
    \textbf{
        Ablation Studies on Different Components in \METHOD{}.
    }
    \textbf{Multi-Rsp.}: Sampling multiple candidate responses before decomposition.
    \textbf{Intra-Topic Rsp}: Intra-topic self-resampling. 
    \textbf{SR}: Selective replacement.
    \textbf{Both/Pref}: Performing selective replacement for both preferred and rejected responses or only for preferred responses while using original responses as rejected counterparts.
    \textbf{In-Ctx}: In-context rewriting for selective replacement.
    Here, we choose \textbf{(2a)} as our baseline, and any difference in other variants are emphasized in \diff{color}.
}
\label{tab:ablation_components}
\begin{center}
\resizebox{\linewidth}{!}{%
\begin{tabular}{cccccccccc}
    \toprule
    & 
    \multicolumn{3}{c}{\textbf{Alternative Generate}} &
    \multicolumn{2}{c}{\textbf{Preference Curation}} &
    \multicolumn{2}{c}{\textbf{ObjHal}} &
    \multicolumn{2}{c}{\textbf{AMBER}} \\
    \cmidrule(lr){2-4}\cmidrule(lr){5-6}\cmidrule(lr){7-8}\cmidrule(lr){9-10} 
    & Multi-Res. & Decompose & Intra-Topic Rsp. & Strategy & In-Ctx. & CH$_\text{s}$ $\downarrow$ & CH$_\text{i}$ $\downarrow$ & Acc. $\uparrow$ & F1 $\uparrow$ \\
    \midrule
    \textbf{(2a)} & \checkmark & \checkmark & \checkmark & SR+Both+Greedy & \checkmark & \textbf{5.9} & \textbf{3.1} & \textbf{82.1} & \textbf{87.0} \\
    \textbf{(2b)} & \checkmark & \checkmark & \checkmark & SR+Both+\diff{Random} & \checkmark & 29.7 & 13.1 & 78.9 & 84.1 \\
    \textbf{(2c)} & \checkmark & \checkmark & \checkmark & SR+\diff{Pref}+Greedy & \checkmark & \underline{6.4} & \underline{3.2} & 76.8 & 84.7 \\
    \textbf{(2d)} & \checkmark & \checkmark & \checkmark & SR+Both+Greedy & \diff{$\usym{2717}$} & 35.5 & 20.1 & 80.1 & 84.4 \\
    \midrule
    \textbf{(2e)} & \diff{$\usym{2717}$} & \checkmark & \checkmark & SR+Both+Greedy & \checkmark & 7.2  & 4.0 & 80.2 & \underline{86.4} \\
    \textbf{(2f)} & \checkmark & \checkmark & \diff{$\usym{2717}$} & SR+Both+Greedy & \checkmark & 12.6 & 6.6 & 77.9 & 84.9 \\
    \midrule
    \textbf{(2g)} & \checkmark & \checkmark & -- & \diff{Ranking} & -- & 9.7  & 4.8 & \underline{80.9} & 85.9 \\
    \textbf{(2h)} & \checkmark & \diff{$\usym{2717}$} & -- & \diff{Ranking} & -- & 25.5 & 12.0 & 73.5 & 82.8 \\
    \textbf{(2i)} & \diff{$\usym{2717}$} & \diff{$\usym{2717}$} & -- & \diff{Rewrite+Pref} & -- & 11.3 & 9.8 & 79.3 & 84.5 \\
    \textbf{(2j)} & \diff{$\usym{2717}$} & \diff{$\usym{2717}$} & -- & \diff{Rewrite+Both} & -- & 15.6 & 12.4 & 78.4 & 83.6 \\
    \bottomrule
\end{tabular}
}
\end{center}
\vspace{-1em}
\end{table}

\subsection{Ablation Studies}
\label{subsec:ablation}

To investigate how \METHOD{} optimizes the reward gap configuration by enabling deliberate control over underlying data characteristics within preference pairs, we conduct a series of ablation studies.
As presented in \Cref{tab:ablation_components}, our investigations focus on addressing the following questions:
\textbf{(1)} How do the core components of \METHOD{} influence the quality and data characteristics of the resulting preference pairs, which directly shape the individual reward gaps?
\textbf{(2)} How does the selective replacement mechanism in \METHOD{} impact final performance of the learned policy, when compared against other preference curation approaches, such as ranking-based or rewriting-based methods?
\textbf{(3)} How to construct a high-quality rejected response $y_l$ in preference pair? Is it more effective to employ external ``black-box'' rewriters common in rewriting-based methods or to utilize model's internally resampled candidates, as \METHOD{} does?
\textbf{(4)} From a broader perspective, what is the data efficiency of preference data curated by \METHOD{}, particularly when comparing its performance with that of other methods across varying data volumes?
For computational efficiency, these ablation studies are conducted using a subset of 8,000 curated preference data instances. 
More ablations are provided in the Appendix~\ref{sec:appendix_more_ablations}, including using different model architectures, using the reference model itself as labeler, and \emph{etc.}

\begin{figure}[t]
\begin{center}
\centerline{\includegraphics[width=\linewidth]{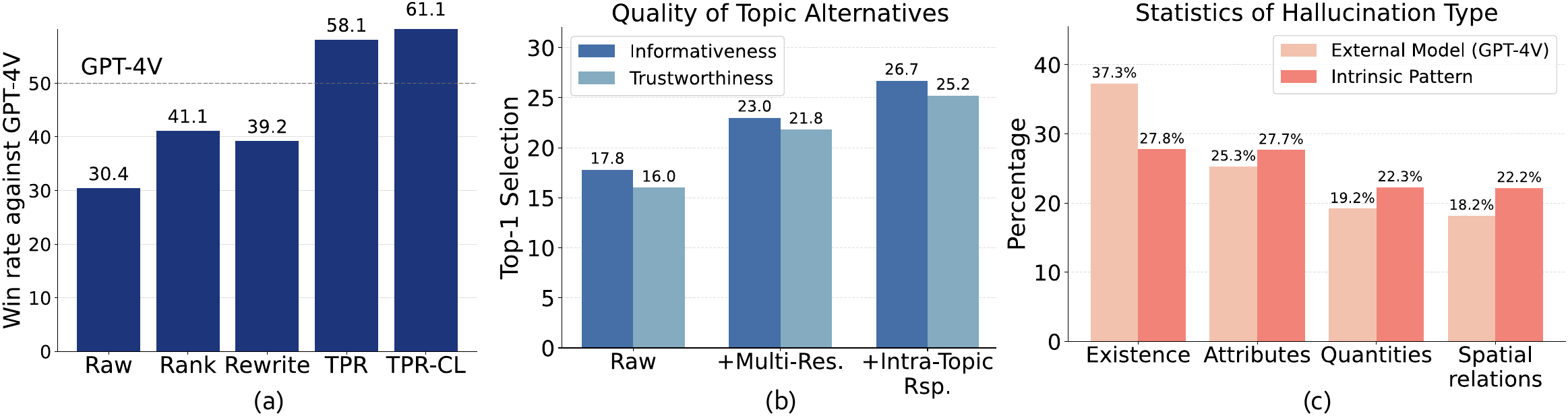}}
\caption{
    \textbf{(a) Quality of Overall Constructed Responses.}
    We compare responses generated from different preference curation strategies against strong ``Ground Truth'' responses from GPT-4V, where win-rates exceeding 50\% indicate superior to GPT-4V outputs.
    For a fair comparison, we use LLaVA-NeXT-34B as the labeler or rewriter across all strategies, including ``+Rank'', ``+Rewrite'' and ``+TPR(-CL)''.
    \textbf{(b) Quality of Topic Alternatives.}
    The numbers on the bars indicate the top-1 selections made by GPT-4V in term of informativeness and trustworthiness.
    \textbf{(c) Hallucination Types.}
    Hallucinations introduced by external rewriters differ significantly from model's own failure modes.
}
\label{fig:ablation_quality}
\end{center}
\vspace{-1em}
\end{figure}

\textbf{Topic Alternatives Generation.}
We conduct ablations on different approaches for generating topic alternatives.
Results in \Cref{tab:ablation_components} reveal that employing multiple initial candidate responses for decomposition and subsequently performing intra-topic self-resampling proves highly beneficial (exp \textbf{2a} \versus{}\,\textbf{2e/f}).
It facilitates better coverage of edge cases, such as rarely mentioned attributes, and ultimately yields a more diverse and comprehensive set of alternatives.
Such diversity is crucial for sculpting more sophisticated reward gap configuration during data curation.
To quantify these aspects, \Cref{fig:ablation_quality} (b) provides a direct comparison, evaluating the underlying data characteristics like informativeness and trustworthiness of these generated alternatives, based on GPT-4V reviews.

\textbf{Preference Curation Approaches.}
We conduct ablations to analyze the impact of specific design choices within the preference curation process, as detailed in \Cref{tab:ablation_components}.
Deviations from our base \METHOD{} variant (exp \textbf{2a}), such as employing random selection for replacement (exp \textbf{2b}), applying selective replacement to only one side of preference pair (exp \textbf{2c}, or omitting in-context rewriting (exp \textbf{2d}), all degrade final model performance.
These findings highlight that the purposeful, symmetric replacement for both preferred and rejected responses, coupled with in-context rewriting, is crucial for crafting reward gaps that provide strong and effective learning signals for DPO.

Moreover, \METHOD{}'s core mechanism of selective topic replacement demonstrably outperforms other preference curation strategies, such as ranking-based or rewriting-based methods (exp \textbf{2a} \versus{}\,\textbf{2g/h/i/j}).
These performance advantages are further substantiated by quality assessments present in \Cref{fig:ablation_quality} (a).
In this evaluation, responses generated by different curation strategies are compared against strong ``ground truth'' (GT) responses from GPT-4V.
We calculate one-vs.-one win-rates to quantify the quality of the generated responses, where win-rates exceeding 50\% indicate that the respective preference curation strategy generates responses superior to those of GPT-4V.
Additionally, more qualitative case studies are provided in Appendix~\ref{sec:appendix_qualitative}, offering an intuitive understanding of the quality improvements achieved by \METHOD{}.
Collectively, all these result confirm that \METHOD{}'s meticulous, topic-focused curation not only leads to better performance but also genuinely enhances the intrinsic quality of the preference data for robust VLM alignment.

\textbf{Rejected Response $y_l$.}
We conduct ablations to assess how different $y_l$ curation approaches influence the nature of the resulting $y_l$, particularly its fidelity in incorporating hallucinations that reflect the base model's genuine failure modes.
Some rewriting-based methods~\cite{zhou2024aligning} employ powerful external models, such as GPT-4V, to modify responses,  intending to create suitably ``negative'' $y_l$ instances.
However, a critical concern is that the hallucination patterns introduced by these external rewriters may deviate from genuine deficiencies of the model that we aim to align.
As illustrated in \Cref{fig:ablation_quality} (c), the distribution of hallucination types in $y_l$ from these methods can differ significantly from model's intrinsic hallucination patterns.
For instance, an external rewriter might introduce a more imbalanced distribution of various hallucination types, while the model's own intrinsic hallucination patterns exhibit a relatively uniform distribution.
This discrepancy results in a mismatch between the curated $y_l$ and model's actual deficiencies, thereby impairing alignment efficiency (exp \textbf{2a} \versus{}\,\textbf{2i} \versus{}\,\textbf{2j}) because the model is not optimally trained against its specific hallucinatory tendencies.
Conversely, \METHOD{} performs intra-topic replacement on model-generated response template, ensuring that $y_l$ more accurately mirror the model's intrinsic failure distribution.

\textbf{Data Efficiency.}
Building on the findings that \METHOD{} yields higher quality topic candidates and overall responses, we further investigate its data efficiency, as illustrated in \Cref{fig:teaser} (b).
While RLHF-V~\cite{yu2024rlhf}, leveraging human annotation, demonstrates notable initial efficiency by achieving low hallucination rates with only 1.4k data, its prohibitive costs inherently limit scalability.
In contrast, \METHOD{}, through its automated data curation process, facilitates a rapid reduction in hallucination levels when scaling from 4k to 20k data, ultimately surpassing human-annotated data in both performance and cost-effectiveness.
Other automated methods generally exhibit inferior data efficiency. 
For instance, RLAIF-V~\cite{yu2024rlaifv}, despite appearing competitive with 22k data volume, actually necessitates multiple rounds of data re-generation and retraining (effectively equivalent to an 88k data processing effort), significantly inflating its computational cost compared to \METHOD{}.
These observations underscore how the enhanced data quality, achieved through \METHOD{}'s systematic optimization of reward gap configuration, directly contributes to superior data efficiency.

\section{Limitations and Future Prospects}
\label{sec:limitation}

Our work presents a promising direction for mitigating hallucinations, yet it also has limitations that open avenues for future research.

\textbf{Complexity of \METHOD{} Pipeline.}
The \METHOD{} involves several sequential steps, including response decomposition, topic clustering, self-resampling, and in-context rewriting. 
While each stage is integral to ensuring high-quality data curation, their combination increases the overall complexity and computational overhead. 
In future work, we are committed to improving the framework's efficiency by simplifying the pipeline, \eg{}, by investigating the necessity of the clustering step, to further enhance its throughput and reduce computational demands.

\textbf{Explore Optimal Data Curation Dimensions.}
While our curriculum learning strategy in \METHOD{}-CL demonstrates one effective way to leverage \METHOD{}'s control capabilities, the potential for optimizing data curation extends further.
Investigating more advanced data curation dimensions, predicated on \METHOD{}'s flexible control over semantic details, remains a rich and promising direction for enhancing VLM robustness against more challenging and complex hallucinations.

\textbf{Hallucinations in Multimodal Reasoning Tasks.}
A critical next step is to extend these mitigation strategies to address hallucinations in complex multimodal reasoning tasks~\cite{openai2024o1,qwen2024qvq,team2025kimi}. 
\METHOD{} is currently effective at reducing perceptual hallucinations, but adapting its topic-level control to handle errors in logical or causal reasoning chains presents a significant and valuable challenge for future exploration.
\section{Conclusion}
\label{sec:conclusion}

In this work, we introduce \METHOD{}, a novel paradigm designed for robust VLM alignment.
By systematically controlling the reward gap configuration through topic-level alternatives generation and selective topic rewriting, \METHOD{} facilitates the construction of high-quality preference pairs with precisely controlled characteristics.
This meticulous data curation is pivotal for generating effective training signals that significantly enhance model performance in mitigating hallucinations. 
Comprehensive experiments demonstrate that \METHOD{}, particularly when augmented with our proposed curriculum learning strategy (\METHOD{}-CL), achieves state-of-the-art performance on several hallucination benchmarks, significantly reducing hallucinatory outputs from the reference model. 
\begin{ack}
This work was supported by National Natural Science Foundation of China (62132001), the Shanghai Artificial Intelligence Laboratory, and the Fundamental Research Funds for the Central Universities.
\end{ack}

\bibliographystyle{abbrvnat}

\clearpage
\newpage
\section*{NeurIPS Paper Checklist}

\begin{enumerate}

\item {\bf Claims}
    \item[] Question: Do the main claims made in the abstract and introduction accurately reflect the paper's contributions and scope?
    \item[] Answer: \answerYes{} 
    \item[] Justification: The claims made in the abstract and instruction can reflect the paper's contributions. And these claims can match with the experimental results in \Cref{subsec:main_results}. 
    \item[] Guidelines:
    \begin{itemize}
        \item The answer NA means that the abstract and introduction do not include the claims made in the paper.
        \item The abstract and/or introduction should clearly state the claims made, including the contributions made in the paper and important assumptions and limitations. A No or NA answer to this question will not be perceived well by the reviewers. 
        \item The claims made should match theoretical and experimental results, and reflect how much the results can be expected to generalize to other settings. 
        \item It is fine to include aspirational goals as motivation as long as it is clear that these goals are not attained by the paper. 
    \end{itemize}

\item {\bf Limitations}
    \item[] Question: Does the paper discuss the limitations of the work performed by the authors?
    \item[] Answer: \answerYes{} 
    \item[] Justification: We have a limitation subsection in \Cref{sec:conclusion}. Besides, we follow the independence assumptions in \cite{yu2024rlaifv} that each decomposed semantic topics exhibit weak correlations within a given response, so we can conduct fine-grained and systematic control over each semantic units in responses (see \Cref{subsec:topic_alter_gen}). And experimental results on several hallucination benchmarks and general capabilities benchmarks prove its robustness. We also provide data efficiency analysis in \Cref{subsec:ablation} and computation cost estimation in Appendix.
    \item[] Guidelines:
    \begin{itemize}
        \item The answer NA means that the paper has no limitation while the answer No means that the paper has limitations, but those are not discussed in the paper. 
        \item The authors are encouraged to create a separate "Limitations" section in their paper.
        \item The paper should point out any strong assumptions and how robust the results are to violations of these assumptions (e.g., independence assumptions, noiseless settings, model well-specification, asymptotic approximations only holding locally). The authors should reflect on how these assumptions might be violated in practice and what the implications would be.
        \item The authors should reflect on the scope of the claims made, e.g., if the approach was only tested on a few datasets or with a few runs. In general, empirical results often depend on implicit assumptions, which should be articulated.
        \item The authors should reflect on the factors that influence the performance of the approach. For example, a facial recognition algorithm may perform poorly when image resolution is low or images are taken in low lighting. Or a speech-to-text system might not be used reliably to provide closed captions for online lectures because it fails to handle technical jargon.
        \item The authors should discuss the computational efficiency of the proposed algorithms and how they scale with dataset size.
        \item If applicable, the authors should discuss possible limitations of their approach to address problems of privacy and fairness.
        \item While the authors might fear that complete honesty about limitations might be used by reviewers as grounds for rejection, a worse outcome might be that reviewers discover limitations that aren't acknowledged in the paper. The authors should use their best judgment and recognize that individual actions in favor of transparency play an important role in developing norms that preserve the integrity of the community. Reviewers will be specifically instructed to not penalize honesty concerning limitations.
    \end{itemize}

\item {\bf Theory assumptions and proofs}
    \item[] Question: For each theoretical result, does the paper provide the full set of assumptions and a complete (and correct) proof?
    \item[] Answer: \answerNA{} 
    \item[] Justification: The paper does not include theoretical results.
    \item[] Guidelines:
    \begin{itemize}
        \item The answer NA means that the paper does not include theoretical results. 
        \item All the theorems, formulas, and proofs in the paper should be numbered and cross-referenced.
        \item All assumptions should be clearly stated or referenced in the statement of any theorems.
        \item The proofs can either appear in the main paper or the supplemental material, but if they appear in the supplemental material, the authors are encouraged to provide a short proof sketch to provide intuition. 
        \item Inversely, any informal proof provided in the core of the paper should be complemented by formal proofs provided in appendix or supplemental material.
        \item Theorems and Lemmas that the proof relies upon should be properly referenced. 
    \end{itemize}

    \item {\bf Experimental result reproducibility}
    \item[] Question: Does the paper fully disclose all the information needed to reproduce the main experimental results of the paper to the extent that it affects the main claims and/or conclusions of the paper (regardless of whether the code and data are provided or not)?
    \item[] Answer: \answerYes{} 
    \item[] Justification: We describe the data curation process of \METHOD{} in \Cref{sec:method}, and provide hyper-parameters and prompts used in Appendix. 
    \item[] Guidelines:
    \begin{itemize}
        \item The answer NA means that the paper does not include experiments.
        \item If the paper includes experiments, a No answer to this question will not be perceived well by the reviewers: Making the paper reproducible is important, regardless of whether the code and data are provided or not.
        \item If the contribution is a dataset and/or model, the authors should describe the steps taken to make their results reproducible or verifiable. 
        \item Depending on the contribution, reproducibility can be accomplished in various ways. For example, if the contribution is a novel architecture, describing the architecture fully might suffice, or if the contribution is a specific model and empirical evaluation, it may be necessary to either make it possible for others to replicate the model with the same dataset, or provide access to the model. In general. releasing code and data is often one good way to accomplish this, but reproducibility can also be provided via detailed instructions for how to replicate the results, access to a hosted model (e.g., in the case of a large language model), releasing of a model checkpoint, or other means that are appropriate to the research performed.
        \item While NeurIPS does not require releasing code, the conference does require all submissions to provide some reasonable avenue for reproducibility, which may depend on the nature of the contribution. For example
        \begin{enumerate}
            \item If the contribution is primarily a new algorithm, the paper should make it clear how to reproduce that algorithm.
            \item If the contribution is primarily a new model architecture, the paper should describe the architecture clearly and fully.
            \item If the contribution is a new model (e.g., a large language model), then there should either be a way to access this model for reproducing the results or a way to reproduce the model (e.g., with an open-source dataset or instructions for how to construct the dataset).
            \item We recognize that reproducibility may be tricky in some cases, in which case authors are welcome to describe the particular way they provide for reproducibility. In the case of closed-source models, it may be that access to the model is limited in some way (e.g., to registered users), but it should be possible for other researchers to have some path to reproducing or verifying the results.
        \end{enumerate}
    \end{itemize}

\item {\bf Open access to data and code}
    \item[] Question: Does the paper provide open access to the data and code, with sufficient instructions to faithfully reproduce the main experimental results, as described in supplemental material?
    \item[] Answer: \answerYes{} 
    \item[] Justification: We release the code and the related instructions anonymously in the supplementary.
    \item[] Guidelines:
    \begin{itemize}
        \item The answer NA means that paper does not include experiments requiring code.
        \item Please see the NeurIPS code and data submission guidelines (\url{https://nips.cc/public/guides/CodeSubmissionPolicy}) for more details.
        \item While we encourage the release of code and data, we understand that this might not be possible, so “No” is an acceptable answer. Papers cannot be rejected simply for not including code, unless this is central to the contribution (e.g., for a new open-source benchmark).
        \item The instructions should contain the exact command and environment needed to run to reproduce the results. See the NeurIPS code and data submission guidelines (\url{https://nips.cc/public/guides/CodeSubmissionPolicy}) for more details.
        \item The authors should provide instructions on data access and preparation, including how to access the raw data, preprocessed data, intermediate data, and generated data, etc.
        \item The authors should provide scripts to reproduce all experimental results for the new proposed method and baselines. If only a subset of experiments are reproducible, they should state which ones are omitted from the script and why.
        \item At submission time, to preserve anonymity, the authors should release anonymized versions (if applicable).
        \item Providing as much information as possible in supplemental material (appended to the paper) is recommended, but including URLs to data and code is permitted.
    \end{itemize}

\item {\bf Experimental setting/details}
    \item[] Question: Does the paper specify all the training and test details (e.g., data splits, hyperparameters, how they were chosen, type of optimizer, etc.) necessary to understand the results?
    \item[] Answer: \answerYes{} 
    \item[] Justification: We provide implementation and evaluation details in \Cref{subsec:exp_setup} and Appendix.
    \item[] Guidelines:
    \begin{itemize}
        \item The answer NA means that the paper does not include experiments.
        \item The experimental setting should be presented in the core of the paper to a level of detail that is necessary to appreciate the results and make sense of them.
        \item The full details can be provided either with the code, in appendix, or as supplemental material.
    \end{itemize}

\item {\bf Experiment statistical significance}
    \item[] Question: Does the paper report error bars suitably and correctly defined or other appropriate information about the statistical significance of the experiments?
    \item[] Answer: \answerNo{} 
    \item[] Justification: We compared a large number of methods in \Cref{tab:main_results}, none of which provided statistical significance. It's infeasible to calculate error bars for all these methods (and some of them are not open sourced). Considering that \METHOD{} has demonstrated the effectiveness on a large number of benchmarks and ablations in \Cref{tab:ablation_components}, we believe it can to some extent prove the stability and robustness of \METHOD{}.
    \item[] Guidelines:
    \begin{itemize}
        \item The answer NA means that the paper does not include experiments.
        \item The authors should answer "Yes" if the results are accompanied by error bars, confidence intervals, or statistical significance tests, at least for the experiments that support the main claims of the paper.
        \item The factors of variability that the error bars are capturing should be clearly stated (for example, train/test split, initialization, random drawing of some parameter, or overall run with given experimental conditions).
        \item The method for calculating the error bars should be explained (closed form formula, call to a library function, bootstrap, etc.)
        \item The assumptions made should be given (e.g., Normally distributed errors).
        \item It should be clear whether the error bar is the standard deviation or the standard error of the mean.
        \item It is OK to report 1-sigma error bars, but one should state it. The authors should preferably report a 2-sigma error bar than state that they have a 96\% CI, if the hypothesis of Normality of errors is not verified.
        \item For asymmetric distributions, the authors should be careful not to show in tables or figures symmetric error bars that would yield results that are out of range (e.g. negative error rates).
        \item If error bars are reported in tables or plots, The authors should explain in the text how they were calculated and reference the corresponding figures or tables in the text.
    \end{itemize}

\item {\bf Experiments compute resources}
    \item[] Question: For each experiment, does the paper provide sufficient information on the computer resources (type of compute workers, memory, time of execution) needed to reproduce the experiments?
    \item[] Answer: \answerYes{} 
    \item[] Justification: We describe it in \Cref{subsec:exp_setup}, and also provide a estimated resources cost in Appendix.
    \item[] Guidelines:
    \begin{itemize}
        \item The answer NA means that the paper does not include experiments.
        \item The paper should indicate the type of compute workers CPU or GPU, internal cluster, or cloud provider, including relevant memory and storage.
        \item The paper should provide the amount of compute required for each of the individual experimental runs as well as estimate the total compute. 
        \item The paper should disclose whether the full research project required more compute than the experiments reported in the paper (e.g., preliminary or failed experiments that didn't make it into the paper). 
    \end{itemize}
    
\item {\bf Code of ethics}
    \item[] Question: Does the research conducted in the paper conform, in every respect, with the NeurIPS Code of Ethics \url{https://neurips.cc/public/EthicsGuidelines}?
    \item[] Answer: \answerYes{} 
    \item[] Justification: We have reviewed the NeurIPS Code of Ethics. The curated preference data in \METHOD{} are sampled from publicly available academic datasets. And most of the visual hallucinations we studied are related to perception errors.
    \item[] Guidelines:
    \begin{itemize}
        \item The answer NA means that the authors have not reviewed the NeurIPS Code of Ethics.
        \item If the authors answer No, they should explain the special circumstances that require a deviation from the Code of Ethics.
        \item The authors should make sure to preserve anonymity (e.g., if there is a special consideration due to laws or regulations in their jurisdiction).
    \end{itemize}

\item {\bf Broader impacts}
    \item[] Question: Does the paper discuss both potential positive societal impacts and negative societal impacts of the work performed?
    \item[] Answer: \answerYes{} 
    \item[] Justification: We discuss broader impacts in Appendix.
    \item[] Guidelines:
    \begin{itemize}
        \item The answer NA means that there is no societal impact of the work performed.
        \item If the authors answer NA or No, they should explain why their work has no societal impact or why the paper does not address societal impact.
        \item Examples of negative societal impacts include potential malicious or unintended uses (e.g., disinformation, generating fake profiles, surveillance), fairness considerations (e.g., deployment of technologies that could make decisions that unfairly impact specific groups), privacy considerations, and security considerations.
        \item The conference expects that many papers will be foundational research and not tied to particular applications, let alone deployments. However, if there is a direct path to any negative applications, the authors should point it out. For example, it is legitimate to point out that an improvement in the quality of generative models could be used to generate deepfakes for disinformation. On the other hand, it is not needed to point out that a generic algorithm for optimizing neural networks could enable people to train models that generate Deepfakes faster.
        \item The authors should consider possible harms that could arise when the technology is being used as intended and functioning correctly, harms that could arise when the technology is being used as intended but gives incorrect results, and harms following from (intentional or unintentional) misuse of the technology.
        \item If there are negative societal impacts, the authors could also discuss possible mitigation strategies (e.g., gated release of models, providing defenses in addition to attacks, mechanisms for monitoring misuse, mechanisms to monitor how a system learns from feedback over time, improving the efficiency and accessibility of ML).
    \end{itemize}
    
\item {\bf Safeguards}
    \item[] Question: Does the paper describe safeguards that have been put in place for responsible release of data or models that have a high risk for misuse (e.g., pretrained language models, image generators, or scraped datasets)?
    \item[] Answer: \answerNA{} 
    \item[] Justification: Our released model are aligned for reducing visual hallucinations.
    \item[] Guidelines:
    \begin{itemize}
        \item The answer NA means that the paper poses no such risks.
        \item Released models that have a high risk for misuse or dual-use should be released with necessary safeguards to allow for controlled use of the model, for example by requiring that users adhere to usage guidelines or restrictions to access the model or implementing safety filters. 
        \item Datasets that have been scraped from the Internet could pose safety risks. The authors should describe how they avoided releasing unsafe images.
        \item We recognize that providing effective safeguards is challenging, and many papers do not require this, but we encourage authors to take this into account and make a best faith effort.
    \end{itemize}

\item {\bf Licenses for existing assets}
    \item[] Question: Are the creators or original owners of assets (e.g., code, data, models), used in the paper, properly credited and are the license and terms of use explicitly mentioned and properly respected?
    \item[] Answer: \answerYes{} 
    \item[] Justification: We cite the original paper related to dataset we used in \METHOD{}, see \Cref{subsec:exp_setup}.
    \item[] Guidelines:
    \begin{itemize}
        \item The answer NA means that the paper does not use existing assets.
        \item The authors should cite the original paper that produced the code package or dataset.
        \item The authors should state which version of the asset is used and, if possible, include a URL.
        \item The name of the license (e.g., CC-BY 4.0) should be included for each asset.
        \item For scraped data from a particular source (e.g., website), the copyright and terms of service of that source should be provided.
        \item If assets are released, the license, copyright information, and terms of use in the package should be provided. For popular datasets, \url{paperswithcode.com/datasets} has curated licenses for some datasets. Their licensing guide can help determine the license of a dataset.
        \item For existing datasets that are re-packaged, both the original license and the license of the derived asset (if it has changed) should be provided.
        \item If this information is not available online, the authors are encouraged to reach out to the asset's creators.
    \end{itemize}

\item {\bf New assets}
    \item[] Question: Are new assets introduced in the paper well documented and is the documentation provided alongside the assets?
    \item[] Answer: \answerYes{} 
    \item[] Justification: We provide the documents related to the released code.
    \item[] Guidelines:
    \begin{itemize}
        \item The answer NA means that the paper does not release new assets.
        \item Researchers should communicate the details of the dataset/code/model as part of their submissions via structured templates. This includes details about training, license, limitations, etc. 
        \item The paper should discuss whether and how consent was obtained from people whose asset is used.
        \item At submission time, remember to anonymize your assets (if applicable). You can either create an anonymized URL or include an anonymized zip file.
    \end{itemize}

\item {\bf Crowdsourcing and research with human subjects}
    \item[] Question: For crowdsourcing experiments and research with human subjects, does the paper include the full text of instructions given to participants and screenshots, if applicable, as well as details about compensation (if any)? 
    \item[] Answer: \answerNA{} 
    \item[] Justification: The paper does not involve crowdsourcing nor research with human subjects.
    \item[] Guidelines:
    \begin{itemize}
        \item The answer NA means that the paper does not involve crowdsourcing nor research with human subjects.
        \item Including this information in the supplemental material is fine, but if the main contribution of the paper involves human subjects, then as much detail as possible should be included in the main paper. 
        \item According to the NeurIPS Code of Ethics, workers involved in data collection, curation, or other labor should be paid at least the minimum wage in the country of the data collector. 
    \end{itemize}

\item {\bf Institutional review board (IRB) approvals or equivalent for research with human subjects}
    \item[] Question: Does the paper describe potential risks incurred by study participants, whether such risks were disclosed to the subjects, and whether Institutional Review Board (IRB) approvals (or an equivalent approval/review based on the requirements of your country or institution) were obtained?
    \item[] Answer: \answerNA{} 
    \item[] Justification: The paper does not involve crowdsourcing nor research with human subjects.
    \item[] Guidelines:
    \begin{itemize}
        \item The answer NA means that the paper does not involve crowdsourcing nor research with human subjects.
        \item Depending on the country in which research is conducted, IRB approval (or equivalent) may be required for any human subjects research. If you obtained IRB approval, you should clearly state this in the paper. 
        \item We recognize that the procedures for this may vary significantly between institutions and locations, and we expect authors to adhere to the NeurIPS Code of Ethics and the guidelines for their institution. 
        \item For initial submissions, do not include any information that would break anonymity (if applicable), such as the institution conducting the review.
    \end{itemize}

\item {\bf Declaration of LLM usage}
    \item[] Question: Does the paper describe the usage of LLMs if it is an important, original, or non-standard component of the core methods in this research? Note that if the LLM is used only for writing, editing, or formatting purposes and does not impact the core methodology, scientific rigorousness, or originality of the research, declaration is not required.
    \item[] Answer: \answerYes{} 
    \item[] Justification: We describe the usage of LLMs in \Cref{sec:method}, and we also provide the prompts in Appendix.
    \item[] Guidelines:
    \begin{itemize}
        \item The answer NA means that the core method development in this research does not involve LLMs as any important, original, or non-standard components.
        \item Please refer to our LLM policy (\url{https://neurips.cc/Conferences/2025/LLM}) for what should or should not be described.
    \end{itemize}

\end{enumerate}

\clearpage
\appendix

\setcounter{figure}{4}
\setcounter{table}{2}

\section{Mathematical Grounding for Reward Gap Optimization}
\label{sec:appendix_math_ground}

In this section, we provide a mathematical justification for the systematic optimization of the reward gap, connecting the data curation strategies introduced in \Cref{sec:method} to the underlying learning dynamics of Direct Preference Optimization (DPO)~\cite{rafailov2024direct}.

\subsection{DPO Loss Function and Gradient Analysis}

The DPO loss function is defined as:
\begin{equation}
\mathcal{L}_\text{DPO}(\pi_\theta; \pi_\text{ref}) = -\mathbb{E}_{(x, y_w, y_l) \sim \mathcal{D}} \left[ \log\sigma\left(\beta \log\frac{\pi_\theta(y_w|x)}{\pi_\text{ref}(y_w|x)} - \beta \log\frac{\pi_\theta(y_l|x)}{\pi_\text{ref}(y_l|x)}\right) \right]
\end{equation}
where $\sigma$ is the sigmoid function and $\beta$ is a hyperparameter controlling the deviation from the reference model $\pi_\text{ref}$. 
For a single preference pair, let $M_{\pi_\theta}$ be the policy model's estimated log-probability ratio (or estimated reward gap):
\begin{equation}
    M_{\pi_\theta} = \log\frac{\pi_\theta(y_w|x)}{\pi_\text{ref}(y_w|x)} - \log\frac{\pi_\theta(y_l|x)}{\pi_\text{ref}(y_l|x)}
\end{equation}
The gradient of the loss with respect to the model parameters $\theta$ is then given by:
\begin{equation}
\label{eq:dpo_grad_appendix}
    \nabla_\theta\mathcal{L}_\text{DPO} = -\beta \cdot \sigma(-\beta M_{\pi_\theta}) \left[ \nabla_\theta\log\pi_\theta(y_w|x) - \nabla_\theta\log\pi_\theta(y_l|x) \right]
\end{equation}
The term $\left[ \nabla_\theta\log\pi_\theta(y_w|x) - \nabla_\theta\log\pi_\theta(y_l|x) \right]$ determines the \textbf{direction} of the update, encouraging the model to increase the likelihood of the preferred response $y_w$ and decrease that of the rejected response $y_l$. 
The term $\sigma(-\beta M_{\pi_\theta})$ determines the \textbf{magnitude} of the gradient update. 
Our analysis focuses on this magnitude term.

\subsection{True Reward Gap in Learning Dynamics}

Let $r^*(y, x)$ represent an oracle or true reward function that perfectly captures the desired behavior (\eg{}, factual accuracy). 
For any preference pair $(y_w, y_l)$ we curate, we can define the \textbf{true reward gap} as:
\begin{equation}
    \Delta r^* = r^*(y_w, x) - r^*(y_l, x)
\end{equation}
This $\Delta r^*$ is an intrinsic property of the data pair that reflects its difficulty. 
A large $\Delta r^*$ signifies an ``easy'' pair (\eg{}, a factually correct response vs. a blatant hallucination), while a small $\Delta r^*$ signifies a ``hard'' pair (\eg{}, a correct response vs. a subtly flawed one).

The curriculum learning strategy in \METHOD{} (see \Cref{subsec:hard_negative_mine}) systematically manipulates $\Delta r^*$ to optimize the learning trajectory. 
We analyze the impact of this manipulation on the model's estimated gap $M_{\pi_\theta}$ and, consequently, the gradient magnitude.

\textbf{Training with High Reward Gaps (Warm-Up Stage).}
Initially, the model is trained on pairs with a large true reward gap ($\Delta r^* \gg 0$). 
The model quickly learns to differentiate these ``easy'' examples, causing its estimated gap $M_{\pi_\theta}$ to become large and positive.
As $M_{\pi_\theta} \to \infty$, the gradient magnitude term $\sigma(-\beta M_{\pi_\theta}) \to 0$. 
The learning signal for these easy examples diminishes, indicating the model has mastered them.

\textbf{Training with Low Reward Gaps (Hard-Mining Stage).}
After the warm-up stage, \METHOD{} introduces ``hard'' pairs where $y_l$ is subtly incorrect, corresponding to a small true reward gap ($\Delta r^* \to 0$). 
For these challenging pairs, the model initially struggles to distinguish $y_w$ from $y_l$, resulting in an estimated gap $M_{\pi_\theta} \approx 0$. 
When $M_{\pi_\theta}$ is near zero, the gradient magnitude $\sigma(-\beta M_{\pi_\theta})$ is at its maximum ($\approx 0.5$). 
This creates a strong learning signal, forcing the model to focus on the fine-grained details it was previously ignoring and refine its decision boundary.
In our curriculum strategy, by progressively reducing the true reward gap in the training data, we keep the model training in a high-gradient magnitude regime for increasingly difficult problems.
This leads to a more robust and fine-tuned policy against challenging hallucinations.

\subsection{Comparison with Adjusting the Hyperparameter $\beta$}

Based on \Cref{eq:dpo_grad_appendix}, an alternative way to amplify the learning signal when $M_{\pi_\theta}$ is large would be to dynamically decrease the hyperparameter $\beta$. 
However, this approach has potential drawbacks. 
The parameter $\beta$ is fundamentally designed to control the strength of the KL-divergence penalty between the policy $\pi_\theta$ and the reference $\pi_\text{ref}$. 
Aggressively lowering $\beta$ to force learning on hard examples could weaken this constraint, potentially leading to unforeseen distribution shifts away from the reference model.
In contrast, \METHOD{}'s approach of manipulating the data's intrinsic reward gap $\Delta r^*$ is more direct. 
By using the reference model itself for every creative step (resampling, rewriting), \METHOD{} ensures that even the ``hard'' preference pairs remain grounded in the model's own failure modes and capabilities, mitigating the risk of drastic policy divergence.

\section{Topic Clustering}
\label{sec:appendix_topic_cluster}

To determine if two semantic units, $u_{m,n}$ and $u_{p,q}$, belong to the same topic $c$, we evaluate their textual consistency and visual correlation.

\textbf{Textual Consistency.}
We assess textual consistency by querying the reference model $\pi_\text{ref}$ to determine if $u_{m,n}$ and $u_{p,q}$ describe the same core idea.
The model is prompted as follows:
\begin{equation}
\label{eq:textual_consistency_appendix}
\begin{aligned}
p_\text{text}(u_{m,n}, u_{p,q})=\pi_\text{ref}(u_{m,n}, u_{p,q}|\;
&\text{\textit{``Are}} \;\ \{u_{m,n}\}\;\ \text{\textit{and}} \;\ \{u_{p,q}\}\;\ \text{\textit{describing}}\\ 
&\;\;\text{\textit{the same topic? Please answer Yes or No.''}}) 
\end{aligned}
\end{equation} 
$p_\text{text}$ is considered \texttt{True} if $\pi_\text{ref}$ outputs Yes, and \texttt{False} otherwise.

\begin{figure}
\begin{minipage}{0.5\linewidth}
    \centering
    \includegraphics[width=\linewidth]{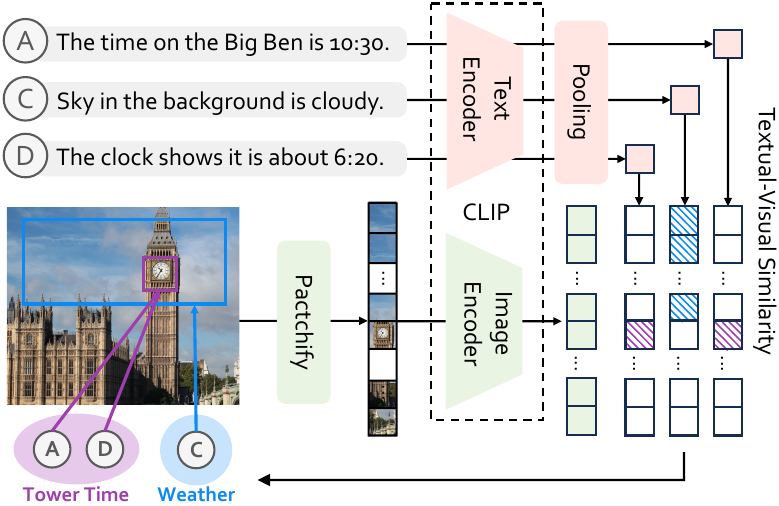}
    \caption{
        \textbf{Visual Correlation.}
    }
    \label{fig:appendix_vis_correlation}
\end{minipage}
\hspace{0.01\linewidth}
\begin{minipage}{0.49\linewidth}
    \centering
    \captionof{table}{
        \textbf{Ablation studies on topic clustering.}
        We ablate the impact of $p_\text{text}$ and $p_\text{vis}$ in topic clustering, and conduct a sensitivity analysis of $\tau_\text{vis}$, used in visual correlation step of topic clustering.
    }
    \resizebox{\linewidth}{!}{
    \begin{tabular}{ccccccc}
    \toprule
    & 
    \multirow{2}{*}{\textbf{Condition}} & 
    \multirow{2}{*}{$\bm{\tau_\text{vis}}$} & 
    \multicolumn{2}{c}{\textbf{ObjHal}} & 
    \multicolumn{2}{c}{\textbf{AMBER}} \\
    \cmidrule(lr){4-5} \cmidrule(lr){6-7}
    & & & 
    CH$_\text{s}$ $\downarrow$ & CH$_\text{i}$ $\downarrow$ & 
    Acc. $\uparrow$ & F1 $\uparrow$ \\
    \midrule
    \textbf{(3a)} & $p_\text{text}$                             & -    & 13.4                 & 6.8             & 80.1             & 85.9             \\
    \textbf{(3b)} & $p_\text{vis}$                            & 0.9  & 13.0                 & 7.3             & 80.1             & 86.1             \\ 
    \midrule
    \textbf{(3c)} & \multirow{4}{*}{$p_\text{text} + p_\text{vis}$} & 0.6  & 8.3                  & 4.1             & 80.8             & 86.1             \\
    \textbf{(3d)} &                                   & 0.8  & 7.2                  & \underline{3.5} & \textbf{82.5}    & \underline{86.9} \\
    \textbf{(3e)} &                                   & 0.9  & \textbf{5.9}         & \textbf{3.1}    & \underline{82.1} & \textbf{87.0}    \\
    \textbf{(3f)} &                                   & 0.95 & \underline{6.7}      & \underline{3.5} & 81.8             & 86.8             \\
    \bottomrule
    \end{tabular}
    }
    \label{tab:appendix_ablation_cluster}
\end{minipage}
\end{figure}

\textbf{Visual Correlation.}
As illustrated in \Cref{fig:appendix_vis_correlation}, we also assess whether two semantic units visually correspond to similar regions within the input image.
To achieve this, we utilize the vision tower in VLM, specifically CLIP~\cite{radford2021learning}, to extract pooled text embeddings for each candidate semantic unit $u$ and image embeddings for each vision token $v$.
We then compute the similarity $\texttt{Sim}(u, v)$ between these text and image embeddings.
Semantic units are considered visually correlated if their respective similarity vectors, when compared to the image embeddings, exhibit a high correlation:
\begin{equation}
\label{eq:visual_correlation_appendix}
p_\text{vis}(u_{m,n}, u_{p,q})=\texttt{Correlation}\Big[\texttt{Sim}(u_{m,n}, v), \texttt{Sim}(u_{p,q}, v)\Big] > \tau_\text{vis}
\end{equation}
Here, $\texttt{Correlation}(\cdot)$, implemented using Pearson correlation, measures the correlation between these two similarity vectors.
$p_\text{vis}$ is considered \texttt{True} if the pre-defined threshold $\tau_\text{vis}$ is satisfied, and \texttt{False} otherwise.
This visual correlation mechanism enables the distinction of multiple entities or aspects within an image, even when their textual descriptions are similar, thereby categorizing them into separate topics.

\textbf{Greedy Clustering.}
Semantic units are considered to belong to the same topic if they exhibit both textual consistency (as defined by $p_\text{text}$) and visual correlation (as defined by $p_\text{vis}$).
Specifically, we adopt an approach analogous to the Louvain method~\cite{blondel2008fast}.
Louvain method is a greedy, iterative algorithm designed to optimize modularity, a metric quantifying the density of connections within communities relative to connections between them.
The process begins by assigning each semantic unit to its own distinct community. 
Subsequently, for each unit, the algorithm evaluates whether moving it to a neighboring community would increase the overall modularity. 
This step is repeated iteratively until no individual move can further enhance modularity. %
This process, consequently, results in a partitioning of the semantic units into distinct topic clusters.

\textbf{Sensitivity Analysis.}
We conducted a sensitivity analysis to evaluate the impact of the hyperparameter $\tau_\text{vis}$ on the final performance.
As detailed in \Cref{tab:appendix_ablation_cluster}, the results indicate that employing a stricter threshold for matching candidate semantic units generally leads to improved performance.
Consequently, based on this analysis, we set $\tau_\text{vis} = 0.9$ for all experimental setups.

\section{Curriculum Learning}
\label{sec:appendix_curriculum_learning}

To progressively refine model's ability to discern subtle errors, \METHOD{}-CL (\textbf{C}urriculum \textbf{L}earning) employs an iterative alignment approach.
It incorporates a selective replacement mechanism for semantic units, guided by a curriculum.
This curriculum dictates the selection of alternative semantic units and subsequently guides the construction of rejected responses $y_l$ paired with preferred ones $y_w$.

\textbf{Implementation Details.}
The \METHOD{}-CL process is divided into multiple iterations. 
In each iteration, preference data is generated using a distinct subset of the overall data sources (as detailed in Section 4.1, containing images $I$ and prompts $x$).
This newly generated preference data is then used to fine-tune the policy model $\pi_\theta$.
The aligned policy model resulting from one iteration serves as the reference model for the subsequent iteration.
Note that unlike methods such as RLAIF-V~\cite{yu2024rlaifv} which re-generates preferences using the entire data sources in each of its multiple iterations (a total generation effort equivalent to $N_\text{iterations}\times |\mathcal{D}|$), \METHOD{}-CL is designed so that preference data for any given subset of the data sources is generated only once throughout the entire multi-iteration process (a total generation effort equivalent to $|\mathcal{D}|$).

During all iterations, preferred responses $y_w$ are constructed by consistently selecting the highest-scored alternative semantic units.
The strategy for constructing rejected responses $y_l$ varies according to the curriculum:
\textbf{(1)} In first 60\% iterations (denoted as ``Warm-Up'' stage), $y_l$ is constructed by replacing semantic units with the lowest-scored alternatives from the candidate pool, encouraging faster initial convergence of the policy model.
\textbf{(2)} In remaining 40\% iterations (denoted as ``Hard-Mining'' stage), the difficulty of the negative examples is gradually increased to further refine the model.
For iterations spanning 60\% to 80\% of the total, alternatives for $y_l$ are selected from those scoring in the bottom 20\% of the candidate pool.
For the final 80\% to 100\% iterations, alternatives for $y_l$ are selected from those scoring between the bottom 20\% and bottom 40\% of the candidate pool.
As the policy model $\pi_\theta$ becomes more capable through these iterations, the progressively increasing score threshold for negative examples challenges it to make finer distinctions.

\begin{figure}
\begin{minipage}{0.55\linewidth}
    \centering
    \includegraphics[width=\linewidth]{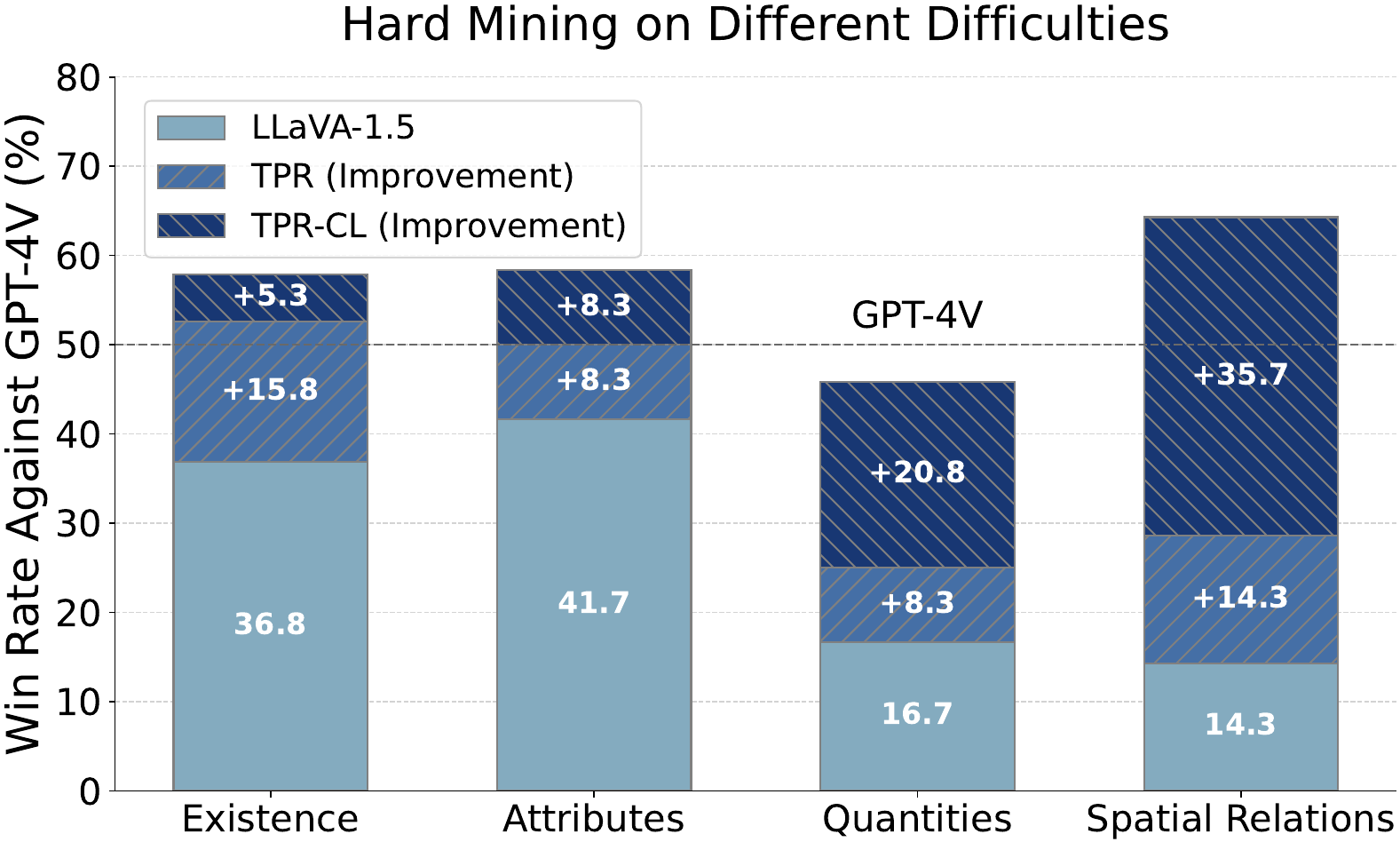}
    \caption{
        \textbf{Effectiveness of Hard Mining.}
        We compare \METHOD{} with curriculum learning strategies against its greedy variant and LLaVA-1.5 baseline.
        Experiments are conducted on RefoMB benchmark using the same training setup for fair comparison.
    }
    \label{fig:appendix_ablation_subtle_hall}
\end{minipage}
\hspace{0.01\linewidth}
\begin{minipage}{0.44\linewidth}
    \centering
    \includegraphics[width=\linewidth]{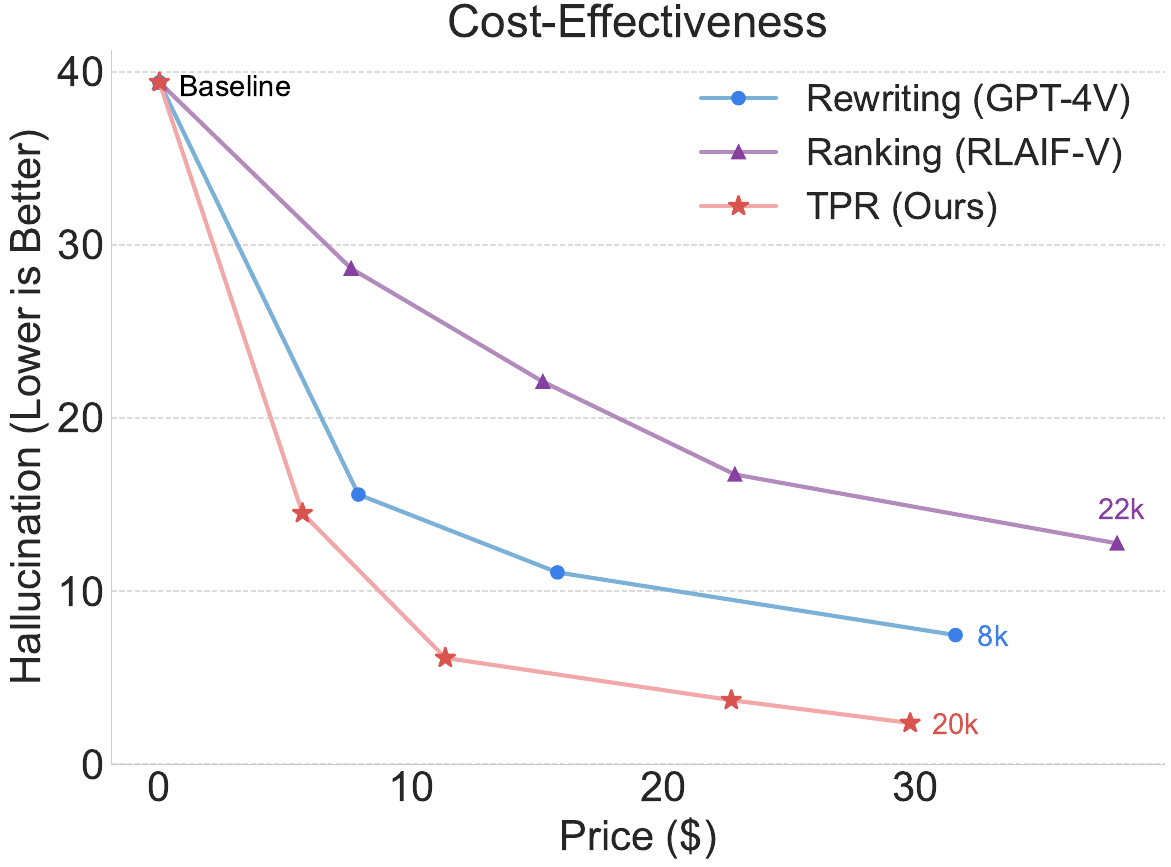}
    \caption{
        \textbf{Cost-Effectiveness.}
        We compare the cost-effectiveness of \METHOD{} against rewriting- and ranking-based methods.
        The numbers next to the lines indicate the total data generation efforts.
    }
    \label{fig:appendix_cost_effectiveness}
\end{minipage}
\end{figure}

\textbf{Effectiveness of Hard Mining.} 
To investigate the impact of curriculum learning on model's ability to avoid subtle errors, we conduct a quantitative analysis comparing effectiveness in reducing hallucination across different types of varying difficulties.
The RefoMB benchmark~\cite{yu2024rlaifv} is used for this ablation, as its detailed categorizations allow us to pinpoint the benefits of curriculum learning more precisely.
The results in \Cref{fig:appendix_ablation_subtle_hall} clearly demonstrate the progressive improvements achieved through our proposed \METHOD{} method and curriculum learning strategy.
On challenging hallucinations for baseline LLaVA-1.5 model, \ie, ``Quantities'' (16.7\%) and ``Spatial Relations'' (14.3\%), the introduction of curriculum learning in \METHOD{}-CL yields further substantial improvements, providing an additional +20.8/+35.7 point increase over its greedy variant.
For ``Existence'' and ``Attributes'', where the greedy variant already performed strongly, \METHOD{}-CL still offers valuable refinements, with additional gains of +5.3 and +8.3 points respectively.
These findings underscore the effectiveness of the hard mining component within our curriculum learning strategy. 
This is especially crucial for tackling more subtle and complex hallucinations related to quantities and spatial relationships, where the baseline model and even the greedy \METHOD{} approach show limitations.

\section{Computational Cost}
\label{sec:appendix_cost}

\begin{figure}
\begin{minipage}{0.51\linewidth}
    \centering
    \captionof{table}{
        \textbf{Computational Cost Breakdown.}
        We provide the total computational cost of \METHOD{} generating full 20k dataset in each stage. 
        vLLM is applied as inference engine in TPR for acceleration.
        All results are obtained using 8 NVIDIA A100 GPUs.
    }
    \resizebox{\linewidth}{!}{
    \begin{tabular}{lcc}
    \toprule
    \textbf{Stage} & \textbf{TPR} & \textbf{TPR w/ vLLM} \\
    \midrule
    Response Generation & 10.27h & 1.36h \\
    Decomposition & 8.91h & 3.57h \\
    Wh-question Convertion & 7.95h & 3.17h \\
    Self-resampling & 7.48h & 2.96h \\
    Topic Cluster & 10.84h & 6.71h \\
    Scoring \& Ranking & 23.43h & 7.42h \\
    In-context Rewriting & 2.17h & 0.85h \\
    \midrule
    \textbf{Total} & \textbf{71.05h} & \textbf{26.04h} \\
    \bottomrule
    \end{tabular}
    }
    \label{tab:appendix_cost_breakdown}
\end{minipage}
\hspace{0.01\linewidth}
\begin{minipage}{0.48\linewidth}
    \centering
    \includegraphics[width=\linewidth]{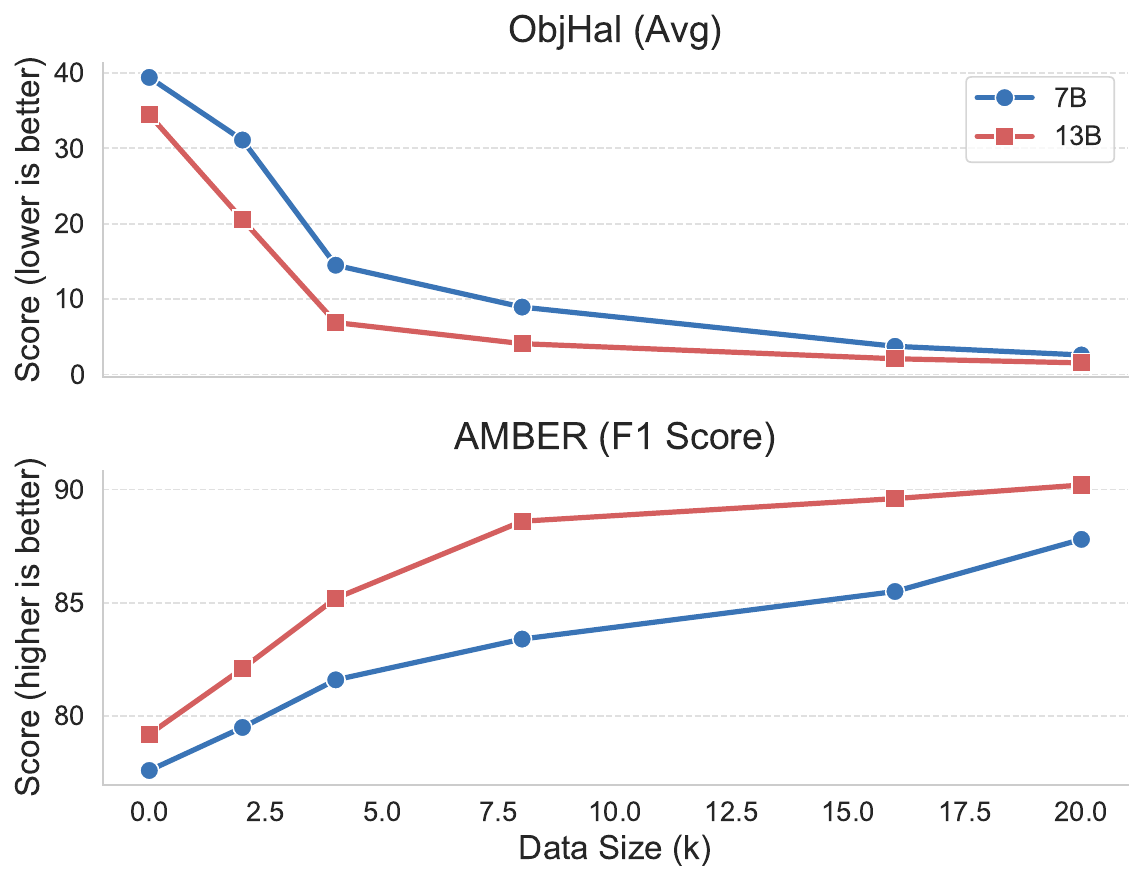}
    \caption{
        \textbf{Larger Base Model.}
        We provide the results of 7B/13B base model on 2k$\sim$20k data.
    }
    \label{fig:appendix_larger_model}
\end{minipage}
\end{figure}

To evaluate the practical viability of \METHOD{}, we conduct a comprehensive computational cost analysis.
As illustrated in \Cref{fig:appendix_cost_effectiveness}, \METHOD{} demonstrates the most favorable efficiency trajectory, achieving a rapid and substantial reduction in hallucinations at a minimal cost.
This superior efficiency is rooted in our data generation pipeline, with a detailed breakdown provided in \Cref{tab:appendix_cost_breakdown}.
The process of generating our 20k preference dataset on 8 NVIDIA A100 GPUs requires 71.05 hours, a time significantly reduced to just 26.04 hours with the integration of the vLLM~\cite{woosuk2023efficient} inference engine. 
This culminates in a highly efficient rate of 4.7 GPU-seconds per generated pair.

When compared with methods relying on proprietary APIs, \METHOD{}'s advantage is pronounced. 
As shown in \Cref{fig:appendix_cost_effectiveness}, while a rewriting approach using GPT-4V yields a significant performance improvement, it comes at a much higher price. 
\METHOD{} achieves a comparable level of hallucination reduction at approximately one-third of the estimated API cost required by the GPT-4V approach (roughly \$10 for \METHOD{} \emph{vs.} over \$30 for GPT-4V).
Furthermore, \METHOD{} is also more cost-effective than other open-source alternatives like RLAIF-V. 
Under the same hardware conditions, generating a 22k dataset with the official RLAIF-V implementation takes approximately 66 hours. 
Although RLAIF-V involves fewer distinct stages, its process is dominated by a computationally expensive scoring phase using a 34B parameter labeler model. 
More critically, its iterative refinement strategy introduces significant overhead: the policy model must be repeatedly retrained between data generation rounds, a resource-intensive process that cannot be accelerated by inference engines like vLLM. 
This inherent inefficiency explains the gentler decline in its performance-to-cost curve shown in \Cref{fig:appendix_cost_effectiveness}.

In summary, these findings clearly indicate that \METHOD{} not only achieves state-of-the-art hallucination reduction but does so with significantly greater computational and financial efficiency than existing methods, highlighting its practical value for large-scale VLM alignment.

\section{Qualitative Case Studies}
\label{sec:appendix_qualitative}

\begin{figure}[h]
	\centering
    \includegraphics[width=\linewidth]{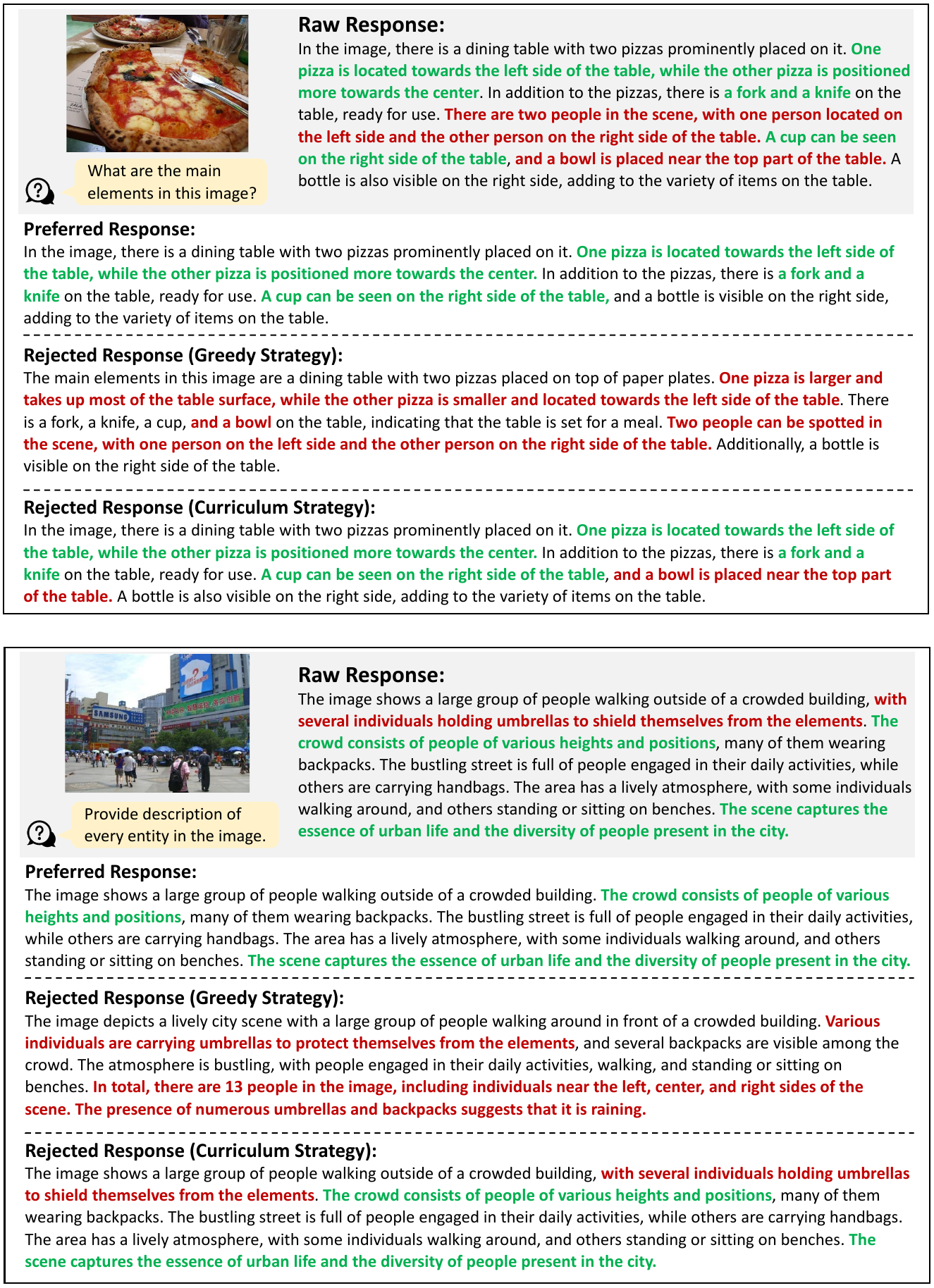}
	\caption{
        \textbf{Qualitative Results of Preferred and Rejected Responses Generated by \METHOD{} and \METHOD{}-CL.}
        \textbf{\textcolor[HTML]{00b050}{Correct answers}} and \textbf{\textcolor[HTML]{c00000}{hallucinations}} are highlighted in color respectively.
    }
	\label{fig:appendix_case_study_curriculum_1}
\end{figure}

\begin{figure}[h]
	\centering
    \includegraphics[width=\linewidth]{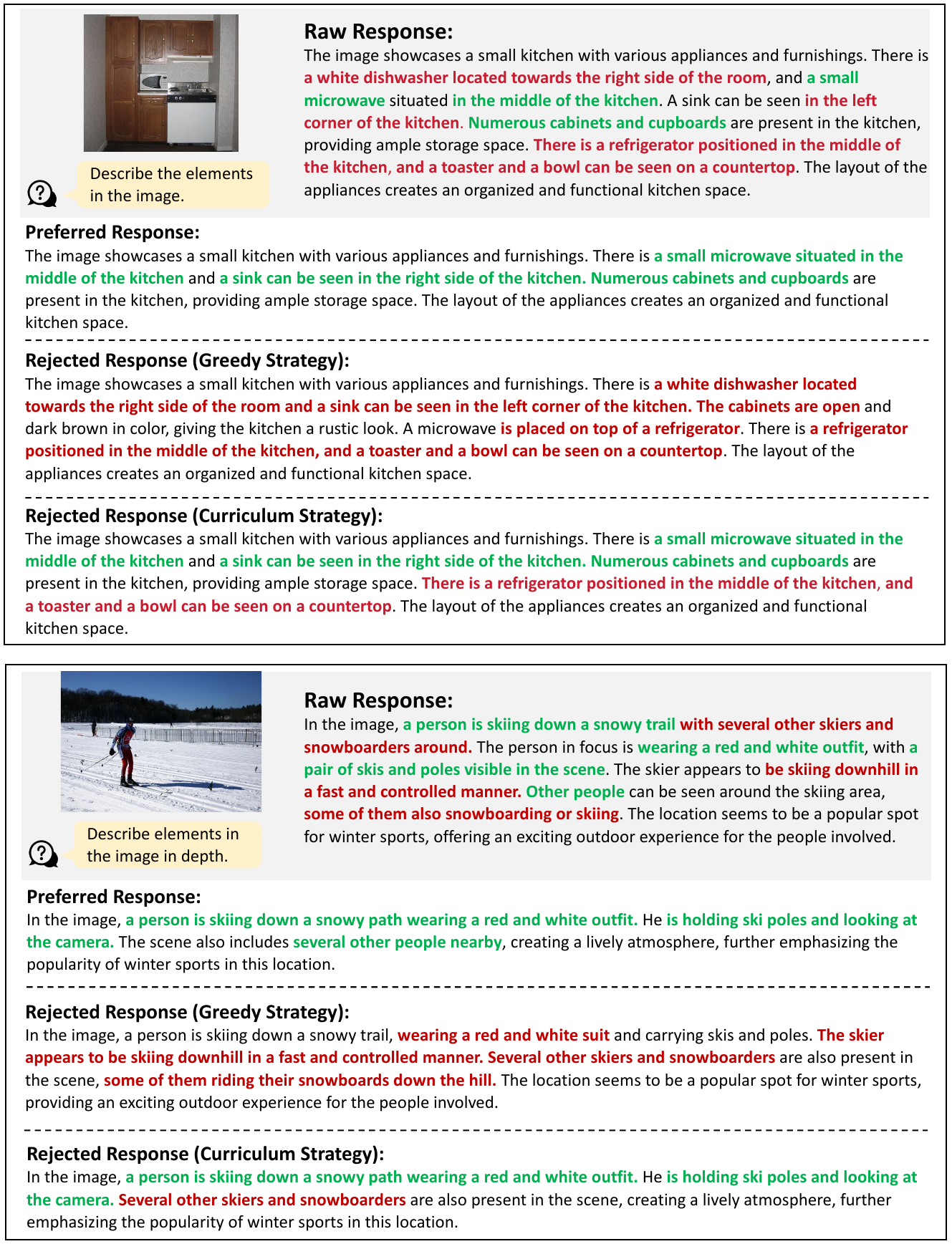}
	\caption{
        \textbf{Qualitative Results of Preferred and Rejected Responses Generated by \METHOD{} and \METHOD{}-CL.}
        \textbf{\textcolor[HTML]{00b050}{Correct answers}} and \textbf{\textcolor[HTML]{c00000}{hallucinations}} are highlighted in color respectively.
    }
	\label{fig:appendix_case_study_curriculum_2}
\end{figure}

\begin{figure}[h]
	\centering
    \includegraphics[width=\linewidth]{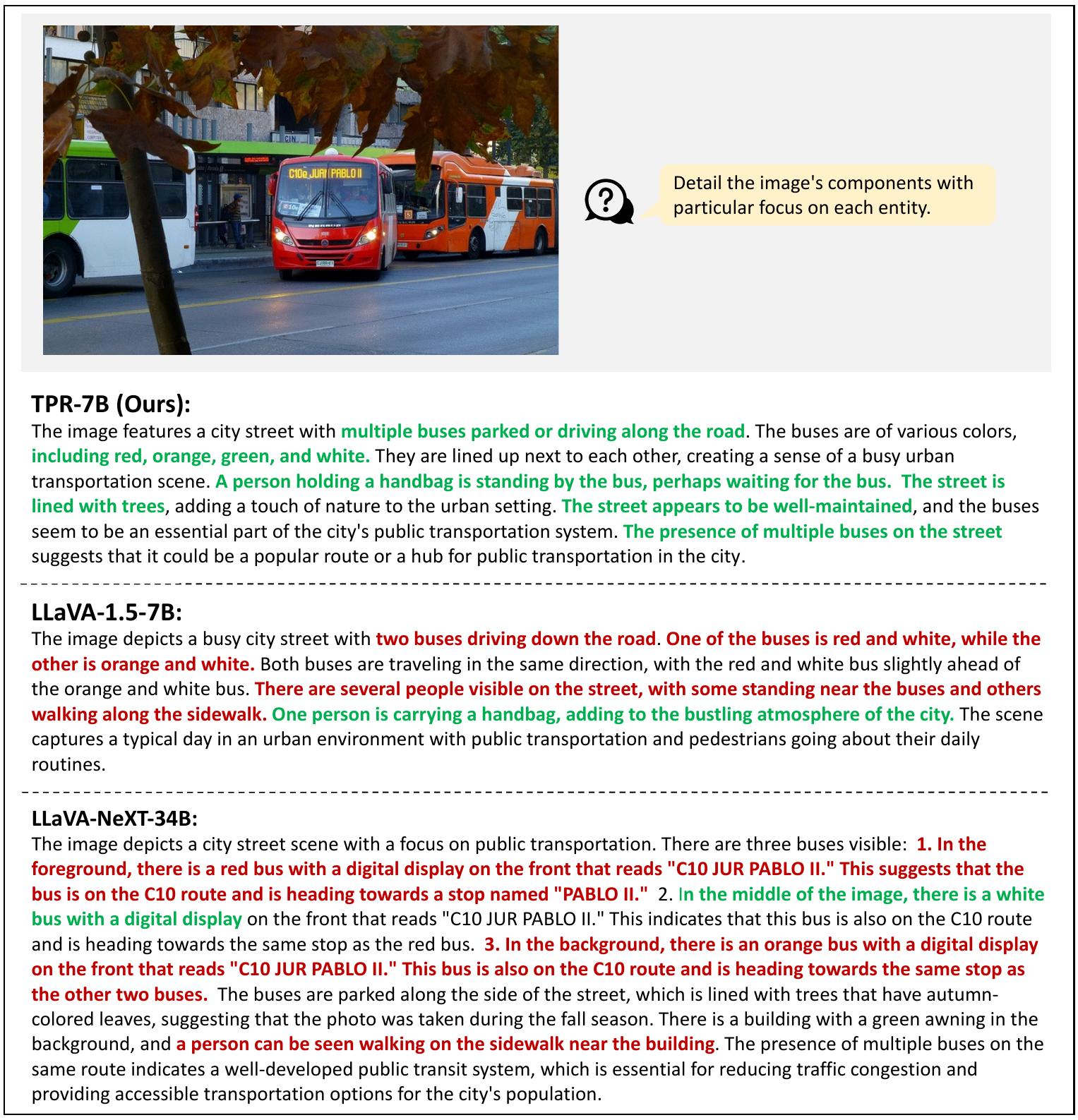}
	\caption{
        \textbf{Qualitative Results of \METHOD{} Compared with LLaVA-1.5-7B and LLaVA-NeXT-34B.}
        \textbf{\textcolor[HTML]{00b050}{Correct answers}} and \textbf{\textcolor[HTML]{c00000}{hallucinations}} are highlighted in color respectively.
    }
	\label{fig:appendix_case_study_aligned_model_1}
\end{figure}

\begin{figure}[h]
	\centering
    \includegraphics[width=\linewidth]{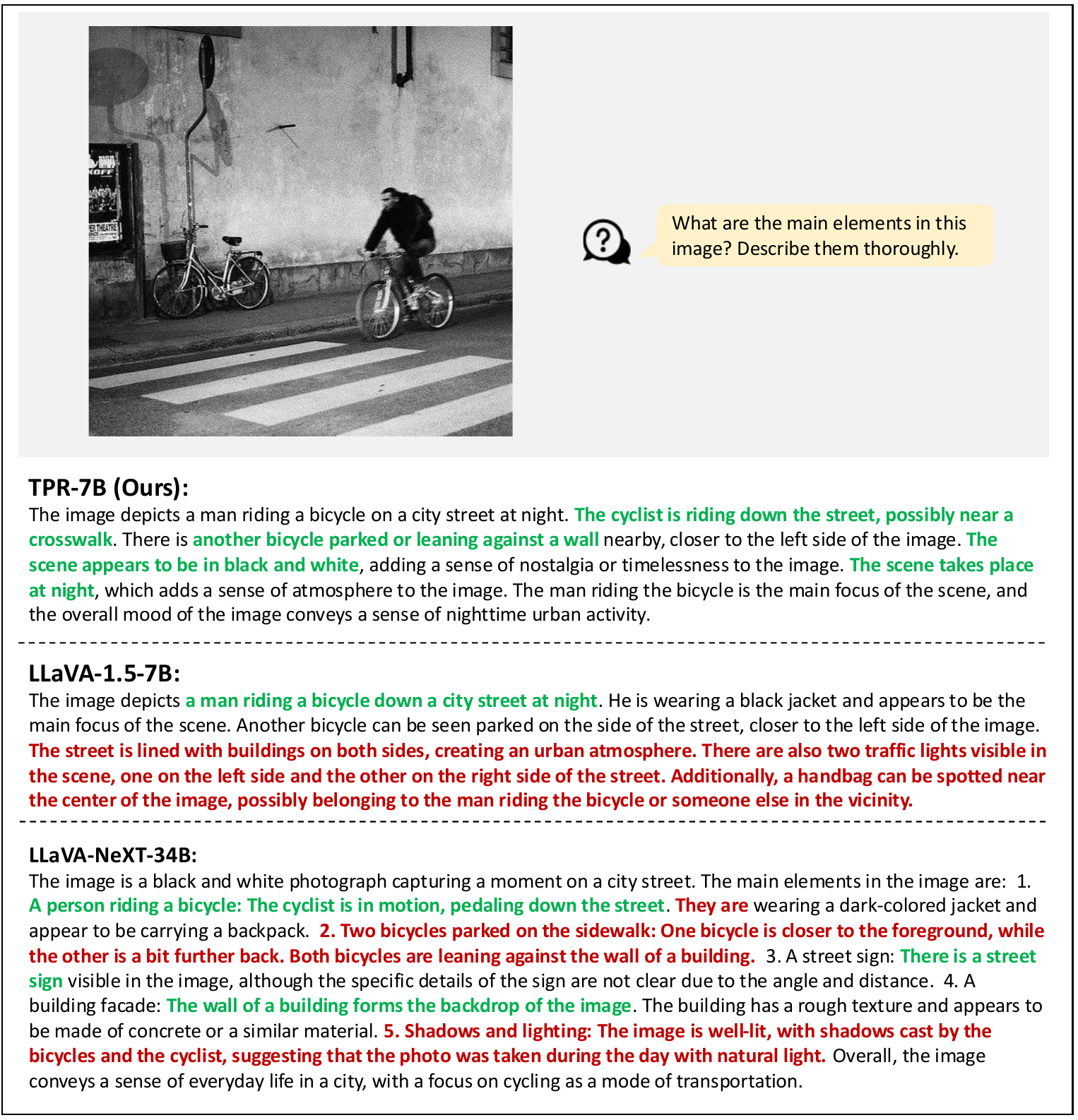}
	\caption{
        \textbf{Qualitative Results of \METHOD{} Compared with LLaVA-1.5-7B and LLaVA-NeXT-34B.}
        \textbf{\textcolor[HTML]{00b050}{Correct answers}} and \textbf{\textcolor[HTML]{c00000}{hallucinations}} are highlighted in color respectively.
    }
	\label{fig:appendix_case_study_aligned_model_2}
\end{figure}

We provide qualitative case studies to illustrate two key aspects: 
\textbf{(1)} The differences between preference pairs generated by \METHOD{} and those generated by its curriculum learning variant, \METHOD{}-CL. See~\Cref{fig:appendix_case_study_curriculum_1} and \ref{fig:appendix_case_study_curriculum_2}.
\textbf{(2)} The enhanced quality of responses from the policy model fine-tuned with \METHOD{}-curated data, particularly when compared against both the LLaVA-1.5-7B baseline and the larger LLaVA-NeXT-34B model. See~\Cref{fig:appendix_case_study_aligned_model_1} and \ref{fig:appendix_case_study_aligned_model_2}.

\section{More Implementation Details}

\subsection{Data Curation}

\textbf{Hyper-Parameters.} 
For preference data curation, we utilize instruction prompt $x$ from RLHF-V~\cite{yu2024rlhf} to guide the reference model $\pi_\text{ref}$ in generating $M=10$ candidate responses $\{y_1,y_2,\cdots,y_M\}$ for each image $I$.
After these responses are decomposed into semantic units, $\pi_\text{ref}$ is queried with wh-questions derived from each unit to self-resample one additional candidate per topic.
For the \METHOD{}-CL variant, 12,000 instances (60\%) are constructed during ``Warm-Up'' stage, and the remaining 8,000 (40\%) are constructed during ``Hard-Mining'' stage.
For initial candidate responses sampling and intra-topic resampling, we follow RLHF-V~\cite{yu2024rlhf} to set the parameters as: \texttt{temperature=0.7,top\_p=0.95,do\_sample=True}. 
Other parameters like \texttt{top\_k} are left at their default values.

\textbf{Prompts.}
Within the \METHOD{} paradigm, the reference model $\pi_\text{ref}$ is prompted to perform several key operations: candidate response generation, response decomposition, topic clustering, in-context rewriting and the conversion of semantic units into both wh-questions and yes-no questions.
The specific prompts utilized to guide $\pi_\text{ref}$ for these tasks are detailed in \Cref{tab:prompt_reference} and \Cref{tab:more_prompt}.

\begin{table}[t]
\centering
\caption{
    \textbf{Prompts Used in Topic-level Alternative Generation.}
    Here, \texttt{\{question\}}, \texttt{\{answer\}} and \texttt{\{sentence\}} are placeholders that will be replaced when inputted into the reference model.
}
\vspace{+4mm}
\begin{minipage}{0.99\linewidth}
\begin{tcolorbox} [colback=white]
    \small

    \textbf{Candidate Response Generation (Modified from RLHF-V)} \vspace{0.1cm} \\
    We randomly select a following question to use as input for the reference model, prompting it to generate responses based on the image.
    
    \vspace{+2mm}
    \textbf{\#\#\# Questions:} \\
    - What is the setting or environment in which the image takes place? \\
    - Provide an intricate description of every entity in the image. \\
    - Can you point out the details that make this image unique? \\
    - What are the main elements in this image? Describe them thoroughly. \\
    - Identify and describe each object in the image in detail. \\
    - Analyze this art image, describing its spatial arrangement, interactive elements, and conceptual message. \\
    - Detail the texts and other components in the image in depth, explaining their relevance to the overall picture. \\
    - Look at the image and describe the celebrity's facial expressions, clothing, and any distinctive features. \\
    - $\cdots$

    \noindent\makebox[\linewidth]{\tikz[baseline]{\draw[dashed] (0,0) -- (\linewidth,0);}}
    
    \vspace{+2mm}
    \textbf{Response Decomposition (Modified from RLAIF-V)} \vspace{0.1cm} \\
    You are an expert in extracting facts from the given question-answer pair for an image.
    Your task is to extract and rewrite the facts mentioned in the answers into self-contained sentences. 
    Exclude opinions or subjective statements.

    \vspace{+2mm}
    You should present your result in the following format:

    \vspace{+1mm}
    \textbf{\#\#\# Facts:} \\
    - \texttt{\{Extracted fact 1\}} \\
    - \texttt{\{Extracted fact 2\}} \\
    - $\cdots$

    \vspace{+1mm}
    \textbf{\#\#\# Question-answer pair:} \\
    Question: \texttt{\{question\}} \\
    Answer: \texttt{\{answer\}}

    \noindent\makebox[\linewidth]{\tikz[baseline]{\draw[dashed] (0,0) -- (\linewidth,0);}}

    \vspace{+2mm}
    \textbf{Wh-Question Converting} \vspace{0.1cm} \\
    You are an expert at modifying a given declarative sentence into a wh-question sentence.
    Your task is to modify the given declarative sentences one by one into a wh-question form. 
    Do not change tenses or add extra content.

    \vspace{+2mm}
    You should present your result in the following format:

    \vspace{+1mm}
    \textbf{\#\#\# Converted questions:} \\
    - \texttt{\{Converted question 1\}} \\
    - \texttt{\{Converted question 2\}} \\
    - $\cdots$

    \vspace{+1mm}
    \textbf{\#\#\# Declarative sentences:} \\
    - \texttt{\{sentence 1\}} \\
    - \texttt{\{sentence 2\}} \\
    - $\cdots$

\end{tcolorbox}
\label{tab:prompt_reference}
\end{minipage}
\end{table}

\begin{table}[t]
\centering
\caption{
    \textbf{Prompts Used in Topic Selective Rewriting.}
    Here, \texttt{\{sentence\}}, \texttt{\{tip\}}, \texttt{\{question\}} and \texttt{\{answer\}} are placeholders that will be replaced when inputted into the reference model.
}
\vspace{+4mm}
\begin{minipage}{0.99\linewidth}
\begin{tcolorbox} [colback=white]
    \small
    
    \textbf{Yes-no Question Converting} \vspace{0.1cm} \\
    You are an expert at modifying a given declarative sentence into a general question sentence. 
    Your task is to modify the given declarative sentences one by one into a general question form. 
    Do not change tenses or add extra content.

    \vspace{+2mm}
    If the given declarative sentence contains not, no or negative meaning words, you need to check the modified general interrogative sentence to make sure that the generated general question sentence retains words with not, no or negative meaning words.

    \vspace{+2mm}
    You should present your result in the following format:

    \vspace{+1mm}
    \#\#\# Converted questions: \\
    - \{Converted question 1\} \\
    - \{Converted question 2\} \\
    - $\cdots$

    \vspace{+1mm}
    \#\#\# Declarative sentences: \\
    - \texttt{\{sentence 1\}} \\
    - \texttt{\{sentence 2\}} \\
    - $\cdots$

    \noindent\makebox[\linewidth]{\tikz[baseline]{\draw[dashed] (0,0) -- (\linewidth,0);}}\vspace{2mm}
    
    \textbf{Textual Consistency Evaluation} \vspace{0.1cm} \\
    You are an expert at determining if the given two declarative sentences are consistent in textual semantics.
    Your task is to determine if the topic described in these two sentences are consistent.
    If you can confirm that two sentences are consistent, please output ``consistent''.
    Otherwise, output ``unrelated''. 

    \vspace{+1mm}
    \#\#\# Declarative sentences: \\
    - \texttt{\{sentence 1\}} \\
    - \texttt{\{sentence 2\}}

    \noindent\makebox[\linewidth]{\tikz[baseline]{\draw[dashed] (0,0) -- (\linewidth,0);}}\vspace{2mm}

    \textbf{In-Context Rewriting} \vspace{0.1cm} \\
    You are an expert at modifying a declarative answer with several tips.
    Your task is to modify the original answer, which is used to answer the question, based on the image and the provided tips.
    The given tips will relate to a specific part of the original answer, and you should use the tips to overwrite the corresponding part.
    If there is a conflict between the tips and the image, remember to follow the tips first.

    \vspace{+2mm}
    You should make minimal modifications and maintain style and format with the original answer. 
    Only output the modified answer.

    \vspace{+1mm}
    \#\#\# Tips: \\
    - \texttt{\{tip 1\}} \\
    - \texttt{\{tip 2\}} \\
    - $\cdots$

    \vspace{+1mm}
    \#\#\# Question-answer pair: \\
    Question: \texttt{\{question\}} \\
    Original Answer: \texttt{\{answer\}}
    
\end{tcolorbox}
\label{tab:more_prompt}
\end{minipage}
\end{table}

\subsection{Evaluation}

\textbf{Benchmarks.}
We evaluate \METHOD{} on several benchmarks:
\begin{itemize}
    \vspace{+0.5mm}
    \item \textbf{Object-HalBench.}
    \citet{rohrbach2018objhal} is designed for common object hallucinations in detailed image descriptions.
    We follow \citet{yu2024rlaifv,yu2024rlhf} to use 8 diverse prompts to improve the stability during evaluation.
    We report the CHAIR$_\text{s}$ (the percentage of hallucinatory responses) and CHAIR$_\text{i}$ (the percentage of hallucinated objects).
    \item \textbf{MMHal-Bench.}
    \citet{sun2024rlhf} evaluates hallucinations and informativeness by using GPT-4~\cite{openai2023gpt4} to compare model outputs with human annotations.
    \item \textbf{AMBER.}
    \citet{wang2023amber} evaluates the object existence, attributes and relations in the image description.
    We use discriminative part of AMBER for evaluation, and report the accuracy and F1 metric.
    \item \textbf{RefoMB.}
    \citet{yu2024rlaifv} consists of 120 images, each paired with 3 instructional annotations, and evaluates 8 fundamental competencies covering both hallucination and reasoning.
    \item \textbf{POPE.}
    \citet{li2023evaluating} evaluates the object existence through querying the VLM with close-ended yes-no questions.
    Note that we use original prompts in POPE during evaluation for stability.
    We report the F1 score and accuracy on three different sampling strategies in POPE, \ie, adversarial, popular and random sampling.
    We also report the overall F1 score.
    \item \textbf{LLaVA-Bench.}
    We use LLaVA-Bench~\cite{liu2023llava} (in-the-wild) to evaluate VLMs in multimodal conversation, detailed descriptions and reasoning aspects.
    We report the overall score.
    \item \textbf{MMStar.}
    \citet{chen2024mmstar} evaluates VLMs on 6 core capabilities and 18 specific aspects related to general capabilities.
    We report the overall score.
\end{itemize}

\textbf{Comparison Counterparts.}
We compare our {\METHOD} with multiple RLHF/RLAIF methods:
\begin{itemize}
    \item \textbf{LLaVA-RLHF.} 
    \citet{sun2024rlhf} first fine-tunes LLaVA~\cite{liu2023llava} with manual-annotated instruction tuning datasets, \ie, VQA-v2~\cite{goyal2017vqav2}, A-OKVQA~\cite{marino2019okvqa} and Flickr30k~\cite{young2014image}, to enhance its general capabilities.
    Subsequently, it trains a reward model on 10k preference data derived from human feedback and applies PPO~\cite{schulman2017proximal} on 72k factually augmented data for preference learning.
    \item \textbf{RLHF-V.}
    \citet{yu2024rlhf} collects 1.4k fine-grained preference data in the form of segment-level corrections on hallucinations through manual annotation.
    It then aligns the VLMs using the proposed dense DPO. 
    \item \textbf{Silkie.}
    \citet{li2024silkie} adopts GPT-4V~\cite{openai2023gpt4v} to assess the responses generated by multiple VLMs regarding helpfulness, visual faithfulness and ethical considerations.
    It then applies DPO~\cite{rafailov2024direct} to train Qwen-VL-Chat~\cite{wang2024qwen} over 80k GPT-4V preferences.
    \item \textbf{POVID.}
    \citet{zhou2024aligning} emphasizes the importance of rejected responses and generates high-quality rejected responses by distorting the image and injecting additional hallucinations using GPT-4V.
    It then fine-tunes the LLaVA-1.5-7B with generated 17k preference data.
    \item \textbf{MFPO.}
    \citet{jiang2024modality} introduces image-related rewards in preference data and constructs 1.4k image preference data upon RLHF-V. 
    It then aligns the VLMs with proposed modality-fair preference optimization (MFPO). 
    \item \textbf{AMP.}
    \citet{zhang2024amp} designs an automated pipeline that generates multi-level preference data for multi-level comparison.
    It then uses 11k multi-level preference data to align VLMs with proposed multi-level DPO.
    \item \textbf{RLAIF-V.}
    \citet{yu2024rlaifv} adopts a divide-and-conquer strategy that determines the overall response score by aggregating the decomposed sub-response scores, mitigating the expensive demand for ultra-large proprietary VLMs.
    It generates 28k preference data for preference learning with proposed iterative DPO.
    \item \textbf{HSA-DPO.}
    \citet{xiao2024detecting} first trains a hallucination detection model on hallucination datasets built by GPT-4V, and then follows a detect-then-rewrite pipeline to construct 6k preference data.
    It then aligns VLMs with proposed hallucination severity-aware DPO.
\end{itemize}
\section{More Ablation Studies}
\label{sec:appendix_more_ablations}

\textbf{Larger Base Model.}
To investigate the scalability of our method, we applied \METHOD{}-CL to both 7B and 13B variants of the base model across data scales from 2k to 20k. 
As illustrated in \Cref{fig:appendix_larger_model}, the 13B model consistently outperforms its 7B counterpart at every data point, achieving better hallucination scores in both ObjHal~\cite{rohrbach2018objhal} and AMBER~\cite{wang2023amber} benchmarks.
This result confirms that the performance gains from \METHOD{} are complementary to the inherent capabilities of larger models, highlighting its strong scalability.

\textbf{Self-Labeling: Using the Reference Model Itself as Labeler.}
The approach of scoring fine-grained semantic units within our intra-topic ranking mechanism potentially alleviates the demand for an exceptionally capable labeler model~\cite{yu2024rlaifv}.
This prompted an exploration into the effectiveness of using the reference model itself for scoring and ranking after self-resampling, thereby investigating the boundaries of model self-improvement.
Accordingly, we introduce \METHOD{}-SL (\textbf{S}elf-\textbf{L}abeling), a variant of \METHOD{} that leverages the base LLaVA-1.5-7B model~\cite{liu2024llava15} as its own preference labeler, instead of a more powerful one like LLaVA-NeXT-34B~\cite{liu2024llavanext} used in our main experiments.
The performance of \METHOD{}-SL compared to the LLaVA-1.5-7B baseline is presented in \Cref{tab:appendix_more_ablation}.
Specifically, \METHOD{}-SL achieves substantial improvements over the baseline LLaVA-1.5-7B across both hallucination mitigation and general VLM capabilities.
These findings are significant as they highlight the effectiveness of \METHOD{} in a self-labeling scenario.
It validates a practical and efficient paradigm for VLM self-improvement that minimizes reliance on human annotations or powerful external models.

\begin{table}[t]
\caption{
    \textbf{More Ablations on Self-Labeling and Different Model Architecture.}
}
\label{tab:appendix_more_ablation}
\vskip 0.15in
\begin{center}
\resizebox{0.98\linewidth}{!}{%
\begin{tabular}{@{}l|ccccccccc|cc}
    \toprule
    & 
    \multicolumn{9}{c|}{\textbf{Hallucination Benchmarks}} & 
    \multicolumn{2}{c}{\textbf{General Benchmarks}} \\
    \cmidrule(lr){2-10} \cmidrule(lr){11-12}
    \multirow{3}{*}{\vspace{+8mm}\textbf{Model}} & 
    \multicolumn{2}{c}{\textbf{ObjHal}} &
    \multicolumn{2}{c}{\textbf{MMHal}} &
    \multicolumn{2}{c}{\textbf{AMBER}} &
    \multicolumn{2}{c}{\textbf{RefoMB}} &
    \multicolumn{1}{c|}{\textbf{POPE}} &
    \multicolumn{1}{c}{\textbf{LLaVA-B}} &
    \multicolumn{1}{c}{\textbf{MMstar}} \\ 
    \cmidrule(lr){2-3} \cmidrule(lr){4-5} \cmidrule(lr){6-7} \cmidrule(lr){8-9} \cmidrule(lr){10-10} \cmidrule(lr){11-11} \cmidrule(lr){12-12}
    &
    CH$_\text{s}$ $\downarrow$ & CH$_\text{i}$ $\downarrow$ &
    Score $\uparrow$ & Hall. $\downarrow$ &
    Acc. $\uparrow$ & F1 $\uparrow$ &
    Trust. $\uparrow$ & Win. $\uparrow$ &
    F1 $\uparrow$ & 
    Overall $\uparrow$ & 
    Overall $\uparrow$ \\
    \midrule
    LLaVA-1.5-7B & 53.6 & 25.2 & 2.36 & 51.0 & 73.5 & 77.6 & 30.8 & 12.1 & 85.9 & 59.7 & 30.3 \\ 
    \;\;+\;\METHOD{}-SL & \textbf{5.8} & \textbf{3.0} & \textbf{2.67} & \textbf{44.8} & \textbf{81.7} & \textbf{86.7} & \textbf{53.5} & \textbf{30.3} & \textbf{86.1} & \textbf{73.7} & \textbf{32.8} \\
    \noalign{\smallskip} \hline \noalign{\smallskip}
    Qwen-VL-2B~\cite{wang2024qwen} & 42.4 & 36.3 & 2.85 & 47.9 & 68.3 & 81.0 & 48.5 & 20.7 & 86.5 & 83.2 & 29.1 \\ 
    \;\;+\;Naive RLAIF     & 36.5 & 31.7 & 2.81 & 49.0 & 70.9 & 81.8 & 52.5 & 23.7 & 87.0  & 83.2 & 30.3 \\
    \;\;+\;\METHOD{}-SL-2B    & \textbf{19.5} & \textbf{13.7} & \textbf{2.98} & \textbf{43.8} & \textbf{74.3} & \textbf{84.3} & \textbf{56.6} & \textbf{28.3} & \textbf{87.1} & \textbf{83.8} & \textbf{31.2} \\
    \bottomrule
\end{tabular}%
}
\end{center}
\vskip -0.1in
\end{table}

\textbf{Generalization on Different Model Architecture.}
To verify the generalizability of the \METHOD{} paradigm, we apply it to a different model architecture, Qwen-VL-2B~\cite{wang2024qwen}.
Specifically, Qwen-VL-2B serves as the reference model for both generating topic-level alternatives and subsequent selective replacement, with intra-topic ranking performed by Qwen-VL-2B itself (self-labeling).
The preference data curated through this process is then used to fine-tune Qwen-VL-2B.
As presented in \Cref{tab:appendix_more_ablation}, we compare this aligned policy model against the original Qwen-VL-2B baseline and the same base model fine-tuned using a na\"ive RLAIF approach (which employs an external LLaVA-NeXT-34B model for ranking).
The results clearly demonstrate the effectiveness and generalizability of applying \METHOD{} paradigm, even in a self-labeling setup, to improve performance on both hallucination and general benchmarks.
This aligns with the conclusions outlined in our main results (see Section 4.2).
Therefore, it is affirmed that the \METHOD{} paradigm is not confined to a specific model architecture like LLaVA.
Its core data curation design can be effectively adapted to other models such as Qwen-VL-2B, leading to significant reductions in hallucinations while maintaining or improving general VLM capabilities.

\section{Broader Impacts}

The research presented in this paper aims to enhance the reliability of Vision Language Models (VLMs). 
By significantly reducing visual hallucinations, \METHOD{} leads to more factually accurate and contextually coherent outputs, which in turn can bolster user trust. 
These improvements are pivotal for unlocking safer and more effective VLM applications across diverse domains, from assistive technologies to educational tools, transforming these models into more reliable instruments.
However, the deployment of this technology require careful consideration of potential societal impacts and technical trade-offs.

\textbf{Potential for Misuse.} 
A primary concern is that the synthetic ``rejected'' responses ($y_l$) generated by \METHOD{}, which are designed to contain plausible hallucinations, could be isolated and misused for creating disinformation if taken out of context.
To mitigate this risk, we are committed to responsible asset release. 
Any public distribution of our curated preference dataset will be governed by a strict, research-focused license prohibiting malicious use. 
Furthermore, the dataset will be accompanied by comprehensive documentation that explicitly warns users that the negative examples are synthetic, generated solely for model alignment, and are factually incorrect and unsuitable for any other application.

\textbf{Environmental and Computational Cost.}
Large-scale data curation and model alignment inevitably have an environmental cost due to the required computational resources. 
As detailed in our analysis (see \Cref{sec:appendix_cost}), \METHOD{} is applied during the post-training alignment phase, making its environmental impact substantially lower than that of pre-training a VLM from scratch.
Crucially, our work emphasizes data efficiency.
By generating high-quality preference pairs that provide strong learning signals, \METHOD{} achieves state-of-the-art performance with less data than many competing methods, thereby promoting a smaller environmental footprint for VLM alignment.

\textbf{Technical Safeguards.} 
Besides, a key technical consideration when applying \METHOD{} is ensuring that targeted improvements in factuality do not compromise the model's broader general capabilities. 
Although our experiments (see \Cref{tab:main_results}) show that our approach maintains or even enhances general performance, achieving this balance requires ongoing research and careful validation.

Successfully navigating these challenges is essential for the responsible development and widespread beneficial adoption of advanced VLM technology.

\end{document}